\documentclass[11pt, a4paper, onecolumn, thu]{thuc3i}

\usepackage{amsmath,amsfonts,bm}

\def\eqref#1{equation~\ref{#1}}
\def\1{\bm{1}}

\DeclareMathAlphabet{\mathsfit}{\encodingdefault}{\sfdefault}{m}{sl}
\SetMathAlphabet{\mathsfit}{bold}{\encodingdefault}{\sfdefault}{bx}{n}

\usepackage{hyperref}
\usepackage{url}
\usepackage{graphicx} 
\usepackage{enumitem}
\usepackage{array}
\usepackage{booktabs}
\usepackage{xcolor}
\usepackage{colortbl}
\usepackage{multirow}
\usepackage{adjustbox}
\usepackage{fontawesome5}
\usepackage{tcolorbox}
\usepackage{verbatim}
\usepackage{float}
\usepackage{graphicx}
\usepackage{amsmath} 
\usepackage{subfig}
\usepackage{tcolorbox}
\tcbuselibrary{skins,breakable}
\usepackage{fontawesome5}
\usepackage{marvosym}
\usepackage[authoryear, sort&compress, round]{natbib}

\definecolor{Gold}{rgb}{1, 0.88, 0.22}
\definecolor{Silver}{rgb}{0.87, 0.87, 0.87}
\definecolor{Bronze}{rgb}{0.88, 0.62, 0.40}

\colorlet{GoldD}{Gold!95!black}
\colorlet{SilverD}{Silver!95!black}
\colorlet{BronzeD}{Bronze!95!black}
\newcommand{\mc}[2]{\textbf{\textcolor{#1D}{#2}}}

\newcommand{\medalbox}[2]{\begingroup\setlength{\fboxsep}{0.5pt}\colorbox{#1}{\strut #2}\endgroup}

\setheadertext{P1 Technical Report}

\title{P1: Mastering Physics Olympiads with Reinforcement Learning}

\author{
Jiacheng Chen$^*$,
Qianjia Cheng$^*$,
Fangchen Yu$^*$,
Haiyuan Wan,
Yuchen Zhang,
Shenghe Zheng, %\quad\quad\quad\quad
Junchi Yao,
Qingyang Zhang,
Haonan He,
Yun Luo,
Yufeng Zhao,
Futing Wang,
Li Sheng,
Chengxing Xie,
Yuxin Zuo,
Yizhuo Li,
Wenxuan Zeng,
Yulun Wu,
Rui Huang,
Dongzhan Zhou,
Kai Chen,
Yu Qiao,
Lei Bai\textsuperscript{\Letter},
Yu Cheng\textsuperscript{\Letter},\quad\quad\quad\quad
Ning Ding\textsuperscript{\Letter},
Bowen Zhou\textsuperscript{\Letter},
Peng Ye\textsuperscript{\Letter$\dagger$},
Ganqu Cui\textsuperscript{\Letter$\dagger$}\\
\vspace{1mm}
\centering{\normalsize P1 Team, Shanghai AI Laboratory}\\
\vspace{2mm}
$^*$ Equal Contribution~~  \textsuperscript{\Letter} Corresponding Authors~~ $^\dagger$ Technical Leads\\
\vspace{1mm}
\faEnvelope[regular]~\texttt{cuiganqu@pjlab.org.cn, yepeng@pjlab.org.cn}  \quad
\faGithub~\href{https://prime-rl.github.io/P1/}{P1 Tech Blog}
}

\begin{abstract}
Recent progress in large language models (LLMs) has moved the frontier from puzzle-solving to science-grade reasoning—the kind needed to tackle problems whose answers must stand against nature, not merely fit a rubric. Physics is the sharpest test of this shift, which binds symbols to reality in a fundamental way, serving as the cornerstone of most modern technologies. In this work, we manage to advance physics research by developing large language models with exceptional physics reasoning capabilities, especially excel at solving Olympiad-level physics problems. We introduce P1, a family of open-source physics reasoning models trained entirely through reinforcement learning (RL). Among them, P1-235B-A22B is the \textbf{first open-source model with Gold-medal performance} at the latest International Physics Olympiad (IPhO 2025), and wins 12 gold medals out of 13 international/regional physics competitions in 2024/2025. P1-30B-A3B also surpasses almost all other open-source models on IPhO 2025, getting a silver medal. Further equipped with an agentic framework PhysicsMinions, P1-235B-A22B+PhysicsMinions achieves overall No.1 on IPhO 2025, and obtains the highest average score over the 13 physics competitions. Besides physics, P1 models also present great performance on other reasoning tasks like math and coding, showing the great generalibility of P1 series.
\end{abstract}

\begin{document}

\maketitle

\begin{figure}[h]
    \centering
    \includegraphics[width=\linewidth]{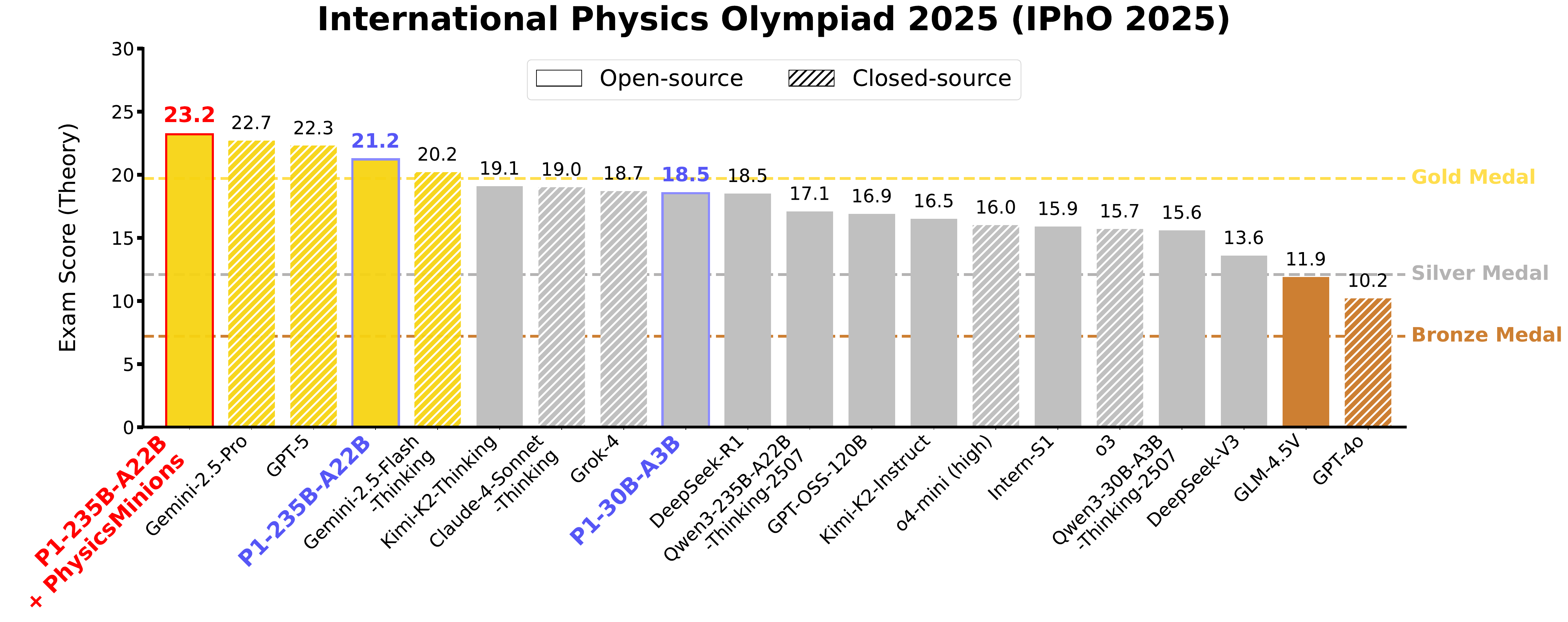}
    \caption{Breakthrough in open-source physics reasoning: \texttt{P1-235B-A22B} stands as the first and only open-source model to win a gold medal at the International Physics Olympiad 2025 (IPhO 2025), placing 3$^{\text{rd}}$ behind Gemini-2.5-Pro and GPT-5. Even at mid-scale, \texttt{P1-30B-A3B} achieved silver and ranked 8$^{\text{th}}$ out of 35 evaluated models, outperforming almost all other open-source models. With the PhysicsMinions agent framework, \texttt{P1-235B-A22B} + PhysicsMinions ranks No.1 on IPhO 2025.”}
    \label{fig:first}
\end{figure}

\newpage
\begingroup
\setlength{\baselineskip}{1.25\baselineskip}
\tableofcontents
\endgroup
\newpage

\section{Introduction}

Recent advances in Large Language Models (LLMs)~\citep{yang2025qwen3,Gemini-2.5,guo2025deepseek-r1} have pushed the boundaries of artificial intelligence from proficiency in symbolic manipulation and general knowledge retrieval to the challenging domain of science-grade reasoning~\citep{gibney2025deepmind_sci,zhang2024llmsci_survey,Intern-S1}. This transition marks a critical frontier: models must no longer simply align with rubrics or datasets, but must produce outputs that withstand the rigorous constraints of physical reality and natural law. Among all scientific disciplines, \textbf{physics} stands as the sharpest and most unforgiving test of this capability—it binds abstract reasoning with empirical consistency, forming the very foundation of modern science and technology.

Mastering physics requires more than factual recall or formulaic application—it demands conceptual understanding, system decomposition, and precise multi-step reasoning grounded in physical law. These skills are most rigorously tested in \textbf{Olympiad-level competitions} such as the International Physics Olympiad (IPhO)~\citep{qiu2025physicsagent,physicsminions}, where each problem condenses the essence of advanced reasoning into a single challenge requiring both analytical precision and creative insight. As such, physics Olympiads provide a high-fidelity and standardized testbed for assessing whether LLMs can exhibit genuine scientific reasoning~\citep{2025hipho}.

We view competitive problem-solving as a critical milestone toward \emph{machine scientific discovery}~\citep{zheng2025llm-discover,oh2025deepmind_discover_rl}. Before models can explore new physical frontiers, they must first demonstrate mastery of human-level reasoning within well-defined laws of nature. Advancing model performance on Olympiad problems is therefore not the end goal, but a necessary step toward building AI systems capable of assisting—or even pioneering—future physics research.

In this work, we introduce \textbf{P1}, a new family of open-source physics reasoning models that push the frontier of scientific AI. Our design integrates both \emph{train-time} and \emph{test-time scaling}, ensuring that the model not only acquires stronger reasoning ability through reinforcement learning (RL)~\citep{sutton1998rl-basis}, but also deploys it adaptively through agentic control at inference.

\begin{itemize}[left=0pt]
    \item \textbf{Train-time Scaling.} 
    The P1 models are trained purely through RL post-training~\citep{guo2025deepseek-r1,cui2025prime-rl} on top of base language models. We propose a multi-stage RL framework that progressively enhances reasoning ability through adaptive learnability adjustment and stabilization mechanisms. This design supports long-term, sustained optimization and effectively mitigates common challenges such as reward sparsity, entropy collapse, and training stagnation.
    \item \textbf{Test-time Scaling.} 
    During inference, we combine P1 models with the \textit{PhysicsMinions}~\citep{physicsminions} agent framework, which equips the model with iterative correction and self-verification abilities. This framework enables multi-turn reflection—allowing P1 to reason, critique, and refine its own solutions, much like human physicists do. Through structured test-time reasoning, the model extends its effective problem-solving depth without additional training.
\end{itemize}

We release two variants of the P1 model family and evaluate them on \textbf{HiPhO}~\citep{2025hipho}, a new benchmark aggregating the latest 13 Olympiad exams from 2024–2025. Our flagship model, \texttt{P1-235B-A22B}, achieves a milestone for the open-source community—becoming the first open model to reach \textbf{Gold-medal performance} on the IPhO 2025, earning 12 golds and 1 silver across the full HiPhO suite. The lightweight variant, \texttt{P1-30B-A3B}, attains \textbf{Silver-medal performance} on IPhO 2025, surpassing almost all prior open baselines. Combined with the PhysicsMinions agent system, P1 attains the \textbf{No.1 overall score} on both the IPhO 2025 and HiPhO leaderboards.

Beyond physics, the P1 models demonstrate remarkable \textbf{generalizability}. The 30B variant significantly outperforms its base model (Qwen3-30B-A3B-Thinking-2507) across seven math, coding, and general reasoning benchmarks, suggesting that physics post-training cultivates transferable reasoning skills rather than domain overfitting.

\paragraph{Contributions.} 
This work makes the following key contributions:
\begin{itemize}[left=0pt]
    \item We introduce \textbf{P1}, the first family of open-source LLMs capable of achieving \textbf{Gold-medal performance} at the International Physics Olympiad level.
    
    \item We develop a \textbf{multi-stage RL post-training framework} with adaptive learnability scaling and training stabilization for sustained reasoning improvement.
    
    \item We unveil the \textbf{P1 ecosystem}—a fully open-source, end-to-end platform encompassing models, training algorithms, an evaluation benchmark, and an agentic inference framework, providing a unified foundation for advancing scientific reasoning in the open community.
\end{itemize}

Together, these advances mark a significant step toward LLMs that can engage in genuine scientific reasoning and eventually contribute to the frontier of physics research.

\section{Physics Dataset}

\subsection{Overview}

\begin{figure}[t]
    \centering 
    \begin{minipage}[c]{0.45\linewidth}
        \centering
        \small 
        \begin{tabular}{l r}
            \toprule
            \textbf{Statistics} & \textbf{Number} \\
            \midrule
            \multicolumn{2}{l}{\textit{Data Composition}} \\
            Total Problems & 5,065 \\
            \quad - Problems from Olympiads & 4,126 (81\%) \\
            \quad - Problems from textbooks & 939 (19\%) \\
            Total Answers & 6,911\\
            Total Fields & 5 \\
            Total Subfields & 25 \\
            Total Answer Types & 6\\
            \midrule
            \multicolumn{2}{l}{\textit{Data Sources}} \\
            Textbooks & 10 \\
            Olympiad Types & 9 \\
            \quad - Sets Collected & 199 \\
            \midrule
            \multicolumn{2}{l}{\textit{Token Statistics}} \\
            Average Question Tokens & 367 \\
            Max Question Tokens & 3386 \\
            Average Solution Tokens & 349 \\
            Max Solution Tokens & 5519 \\
            \bottomrule
        \end{tabular}
        \captionof{table}{Statistics of the training data.}
        \label{tab:data_statistics}
    \end{minipage}
    \hfill 
    \begin{minipage}[c]{0.5\linewidth}
        \centering
        \includegraphics[width=\linewidth]{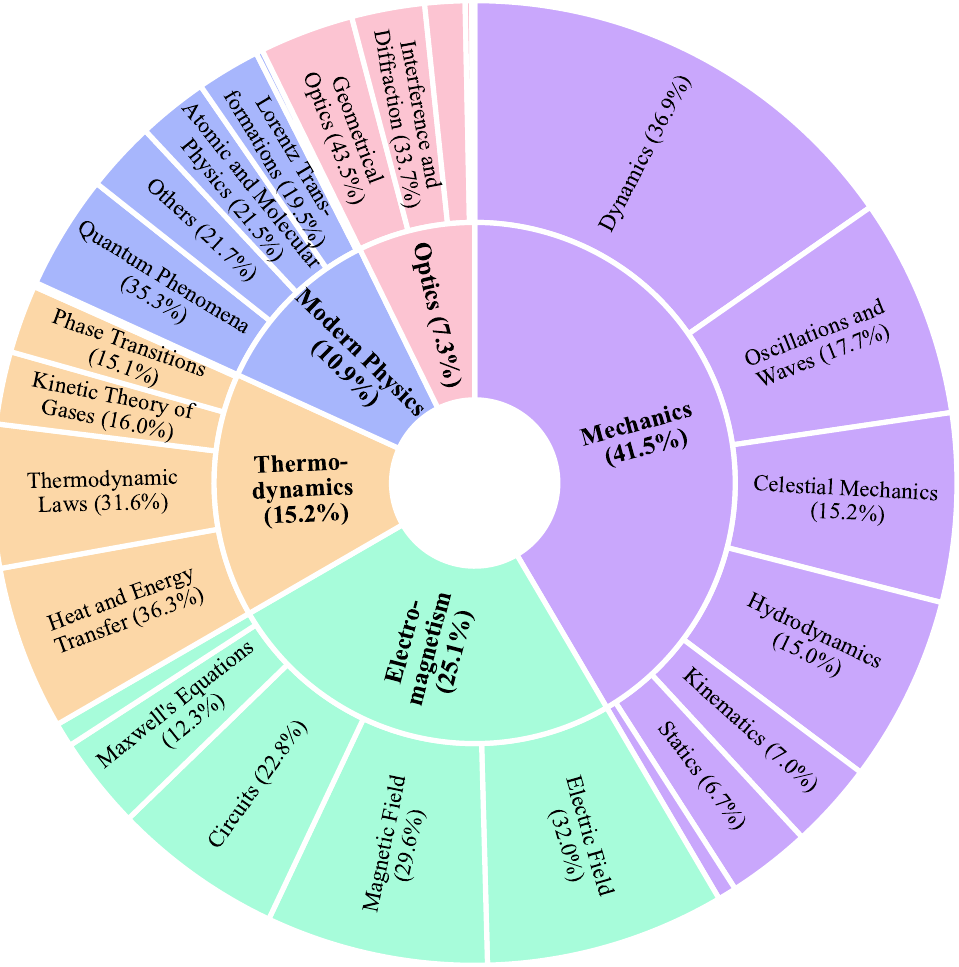}
        \caption{Field distribution of the training data.}
        \label{fig:field_distribution}
    \end{minipage}

\end{figure}

We introduce a systematically curated dataset of 5,065 Olympiad-level, text-based physics problems, designed to advance LLMs toward genuine scientific reasoning. As summarized in Table~\ref{tab:data_statistics}, the dataset combines 4,126 problems from physics Olympiads with 939 from competition textbooks, spanning 5 fields and 25 subfields (see Figure~\ref{fig:field_distribution}). Rather than pursuing broader coverage, we focus on depth and rigor—physics Olympiads uniquely couple abstract reasoning with empirical law, providing an exacting testbed for improving and evaluating model reasoning under physical constraints. Following PHYSICS~\citep{zheng2025scaling} and HiPhO~\citep{2025hipho}, we further refine the data construction pipeline by integrating the strengths of human and model annotation, achieving finer-grained extraction and higher data quality. This dataset thus bridges the gap between general reasoning corpora and science-grade problem solving, offering high-difficulty, high-fidelity supervision for post-training.

\begin{figure}[h]
    \centering
    \includegraphics[width=\linewidth]{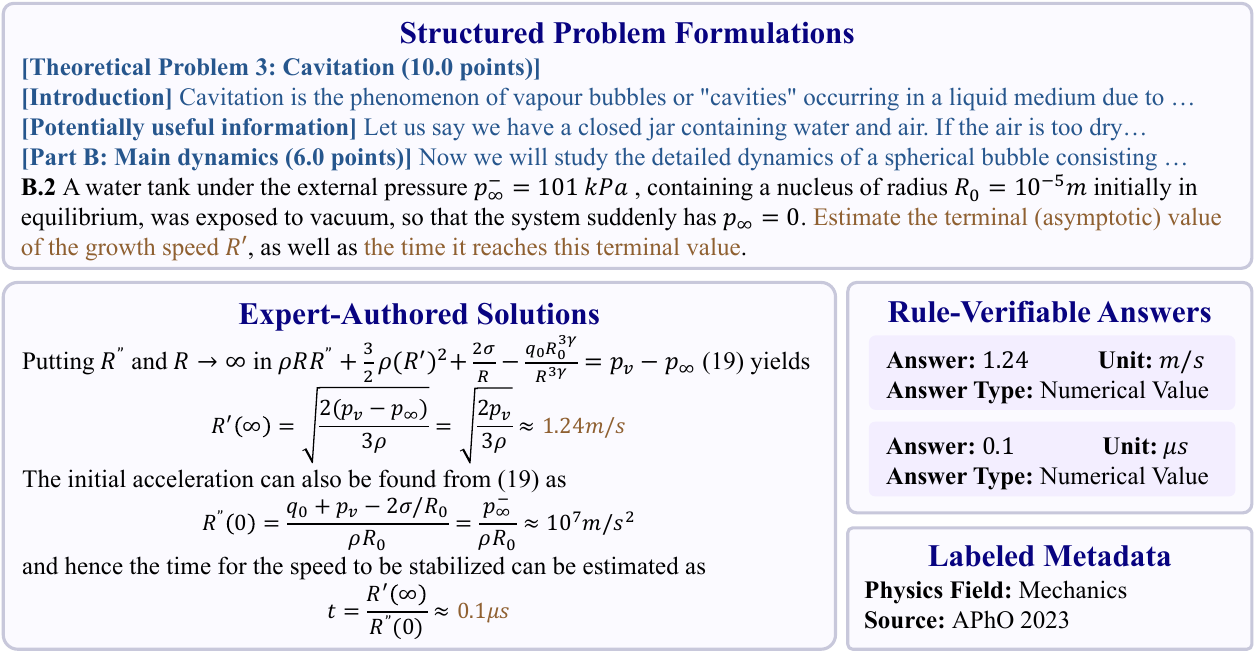}
    \caption{A data sample from the training data. }
    \label{fig:data_example}
\end{figure}

As illustrated in Figure~\ref{fig:data_example}, each instance in the dataset follows a structured \textit{Question--Solution--Answer} schema, enriched with metadata, providing a well-organized format to support diverse avenues of research.

\begin{itemize}[left=0pt, parsep=0pt]
    \item \textbf{Structured Problem Formulations.} Physical problem statements are preserved in their original form whenever possible. For excessively long problems, subdivisions are introduced to respect model context limits while retaining the logical integrity of the task.
    \item \textbf{Expert-Authored Solutions.} Solution processes are authored by human physics experts, providing authentic reasoning trajectories.
    \item \textbf{Rule-Verifiable Answers.} Verifiable final answers supply the unambiguous correctness criteria required for RLVR. Annotations on type, unit, and scoring points support reliable validation and mirror the weighted criteria of human grading.
    \item \textbf{Labeled Metadata.} Each instance is tagged with physics field and source, enabling analysis of domain coverage, guiding data selection strategies, and providing a basis for studying the impact of provenance on training dynamics.
\end{itemize}

\subsection{Dataset Construction}

\paragraph{Data Collection.}
Our data construction is guided by a central objective: enabling LLMs to internalize structured reasoning that aligns with physical laws and empirical consistency—a prerequisite for genuine scientific intelligence. We therefore prioritize problems that combine reasoning depth with rule-verifiable outcomes—conditions empirically linked to stronger reasoning gains in recent studies \citep{guo2025deepseek-r1,yang2025qwen3,wen2025reinforcement}. Within physics, these properties converge most naturally in high school Olympiad problems: they demand rigorous modeling, symbolic precision, and multi-step inference, yet remain tractable without domain-specific obscurity. This has been empirically validated in the PHYSICS~\citep{zheng2025scaling}, where LLMs perform worse on high school Olympiad problems than on undergraduate non-physics-major questions, highlighting the distinct reasoning challenge posed by Olympiad problems. Accordingly, we assemble two complementary sources. The first comprises ten major physics Olympiads(up to 2023)—including APhO, IPhO and others—spanning regional to international tiers and capturing a naturally graded difficulty spectrum. The second consists of ten authoritative competition textbooks, offering systematically organized examples and exercises with expert-authored solutions.

\paragraph{Data Annotation.}
Driven both by the demand for high-quality extraction and the heterogeneity of the sources, the construction process is organized as a multi-stage pipeline. 

\begin{enumerate}[left=0pt]
    \item \textbf{PDF-to-Markdown Conversion.} Source materials in PDF format are parsed into Markdown using Optical Character Recognition (OCR) tools. 
    \item \textbf{Questions and Solutions Parsing.} Extraction strategies are tailored to the two source types. For textbooks, model-assisted parsing leverages structural cues (e.g., chapter boundaries and numbering) to align exercises with their solutions. Olympiad problems, characterized by lengthy statements and multiple sub-questions, are manually restructured by experts to separate shared background from sub-questions, preserving both readability and fidelity. 
    \item \textbf{Answer Annotation.} Answers are automatically extracted by models and decomposed into structured lists, allowing each sub-answer to be individually validated against model outputs. Units are separated into explicit fields to support standardized, rule-based scoring.
    \item \textbf{Language Normalization.} Problems originating in Chinese (e.g., CPhO) are translated into English with Claude to maintain a consistent monolingual corpus.
\end{enumerate}

\paragraph{Quality Control.} To ensure reliability, we implement a multi-stage quality control pipeline, integrating model-based automated procedures and expert review. \textbf{(1) OCR Correction.} Texts parsed from complex page layouts or low-quality scans are manually validated against the original PDFs to correct OCR artifacts. \textbf{(2) Answer Cross-Validation. } Three models—Gemini-2.5-Flash, Claude-3.7-Sonnet, and GPT-4o—independently extract answers from each question–solution pair; Consensus is established when at least two models agree; items without such agreement are removed. \textbf{(3) Data Filtering. } Problems requiring diagram drawing or involving unverifiable answers (e.g., proofs or explanations) are removed. \textbf{(4) Expert Review. } Claude-3.7-Sonnet performs a comprehensive consistency audit, followed by targeted manual refinement. After these stages, the dataset contracts from 6,516 to 5,065 items, yielding an English, text-only corpus with rule-verifiable answers, well-suited for RLVR.
\section{Approach}
\subsection{RL Formulation}

We formulate the problem of solving physics Olympiad tasks as a reinforcement learning (RL)~\citep{sutton1998rl-basis} process. 
Let $\mathcal{M} = (\mathcal{S}, \mathcal{A}, P, r)$ denote the underlying Markov Decision Process (MDP), where:
\begin{itemize}[left=0pt]
    \item $\mathcal{S}$ represents the state space, corresponding to the model context, including the problem statement and all previously generated reasoning steps.
    \item $\mathcal{A}$ is the action space, defined over the token vocabulary from which the model generates its next output token.
    \item $P(s' \mid s, a)$ is the (deterministic) transition function, which appends the newly generated token $a$ to the state $s$, resulting in an updated context $s'$.
    \item $r(s, a)$ is the reward function, which evaluates the correctness and quality of the final solution trace. 
\end{itemize}

The learning objective is to maximize the expected return:
\begin{equation}
    J(\pi_\theta) = \mathbb{E}_{\tau \sim \pi_\theta}\Bigg[\sum_{t=0}^{T} r(s_t, a_t)\Bigg],
\end{equation}
where $\pi_\theta$ is the policy parameterized by model parameters $\theta$, and $\tau = (s_0, a_0, \dots, s_T)$ denotes a trajectory sampled from $\pi_\theta$.

\paragraph{Policy Gradient.}  
The policy gradient~\citep{sutton1999pg-sutton} method optimizes $\pi_\theta$ by ascending the gradient of the expected return:
\begin{equation}\label{eq:pg_loss}
    \nabla_\theta J(\pi_\theta) = \mathbb{E}_{\tau \sim \pi_\theta}\Bigg[\sum_{t=0}^{T} \nabla_\theta \log \pi_\theta(a_t \mid s_t) \, A^\pi(s_t, a_t)\Bigg],
\end{equation}
where $A^\pi(s_t,a_t)$ is the advantage function estimating the relative value of action $a_t$ in state $s_t$.  
We adopt this standard form and instantiate it via GSPO below.

\textbf{Group Sequence Policy Optimization (GSPO).} 
GSPO~\citep{zheng2025gspo} elevates optimization from the token level~\citep{shao2024grpo,yu2025dapo} to the sequence level, employing length-normalized sequence likelihood importance ratios:
\begin{equation}
    s_i(\theta) = \left(\frac{\pi_\theta(y_i|x)}{\pi_{\theta_{\text{old}}}(y_i|x)}\right)^{1/|y_i|} = \exp\left(\frac{1}{|y_i|}\sum_{t=1}^{|y_i|} \log \frac{\pi_\theta(y_{i,t}|x, y_{i,<t})}{\pi_{\theta_{\text{old}}}(y_{i,t}|x, y_{i,<t})}\right),
\end{equation}
where $|y_i|$ denotes the sequence length, and the $1/|y_i|$ term implements length normalization to reduce variance. The corresponding advantage function is computed at the sequence level:
\begin{equation}
    \hat{A}_i^{\text{GSPO}} = R_i - \frac{1}{G}\sum_{j=1}^G R_j,
\end{equation}
with the objective function:
\begin{equation}
    J^{\text{GSPO}}(\theta) = \mathbb{E}_{x \sim \mathcal{D}, \{y_i\}_{i=1}^G \sim \pi_{\theta_{\text{old}}}(\cdot|x)}\left[\frac{1}{G}\sum_{i=1}^G \min\left(s_i(\theta)\hat{A}_i^{\text{GSPO}}, \text{clip}(s_i(\theta), 1-\epsilon, 1+\epsilon)\hat{A}_i^{\text{GSPO}}\right)\right].
\end{equation}

\subsection{Instantiation}

\paragraph{Reward Design. }To instantiate the RL algorithms for solving physics problems, we must concretely define the reward function.  
Following the Correct-or-Not design in RLVR methods~\citep{guo2025deepseek-r1,zheng2025gspo}, we employ a binary reward scheme based on answer correctness, leveraging the fact that our physics dataset contains verifiable ground-truth outcomes:
\begin{equation}
    r = \begin{cases}
        1, & \text{if the predicted answer matches the ground truth}, \\
        0, & \text{otherwise}.
    \end{cases}
\end{equation}

However, physics problems often involve multiple sub-questions or require multiple final results (e.g., solving for both $a$ and $b$).  
To account for this structure, we adopt a test-case-style reward aggregation similar to program evaluation, defining the final reward as:
\begin{equation}
    R = \frac{1}{N}\sum_{i=1}^N r_i,
\end{equation}
where $N$ is the number of required sub-answers in the problem, and $r_i$ denotes the correctness indicator for the $i$-th sub-answer.

\paragraph{Answer Extraction. }We design prompts that enforce a \texttt{multi-box} output format, requiring the model to place each sub-answer sequentially inside separate \texttt{\textbackslash boxed} environments, in order to simplify the answer extraction. Specifically, we apply system prompt as shown in Figure~\ref{fig:system_prompt} into our physics dataset. 

\begin{figure}[h]
\begin{tcolorbox}[colback=gray!5!white, colframe=gray!75!black, title=System Prompt of Multi-box Style]
\begin{verbatim}
Please answer the problem adhering to the following rules:
1. Please use LaTeX format to represent the variables and formulas 
   used in the solution process and results.
2. Please put the final answer(s) in \boxed{}, note that the unit 
   of the answer should not be included in \boxed{}.
3. If the problem requires multiple answers, list them in order, 
   each in a separate \boxed{}.
\end{verbatim}
\end{tcolorbox}
\caption{System prompt design for P1 training.}
\label{fig:system_prompt}

\end{figure}

\paragraph{Verifier Design.}  
To handle the inherent complexity of physics answers, which often appear as symbolic expressions rather than single numeric values, we adopt a hybrid verification framework that integrates both rule-based and model-based components:
\begin{itemize}[left=0pt]
    \item \emph{Rule-based verifier.}  
    Inspired by DrGRPO~\citep{liu2025drgrpo}, we combine symbolic computation with rule-based checks using SymPy~\citep{meurer2017sympy} and math-verify~\citep{math-verify} heuristics. This allows robust equivalence testing of algebraic expressions, including commutativity, factorization, and simplification.
    
    \item \emph{Model-based verifier (used only in validation).}  
    Complementing the rule-based system, we follow the XVerify~\citep{chen2025xverify} paradigm and employ a large language model (Qwen3-30B-A3B-Instruct-2507) as an answer-level verifier. Given the problem statement, the extracted model prediction, and the ground truth, the verifier outputs a binary judgment (\emph{correct} or \emph{incorrect}), improving robustness against cases that are challenging for purely symbolic methods.
\end{itemize}

\subsection{Technical Design}

\subsubsection{Adaptive Learnability Adjustment}

Achieving sustained performance growth during post-training is a key challenge for large language models. In practice, RL fine-tuning often faces performance bottlenecks after an initial phase of rapid improvement~\citep{yu2025dapo,cui2025entropy}. These plateaus can be attributed to entropy collapse, sparse rewards, limited base model capacity, or imperfect data quality~\citep{cui2025entropy,zhang2025survey-reasoning}. We collectively refer to these factors as a reduction in \emph{learnability}. For example, entropy collapse diminishes exploration capability and thus reduces learnability; similarly, if the model capability and the dataset difficulty are mismatched, the model either fails to learn or trivially solves problems without meaningful updates, again resulting in lowered learnability. To address this issue, especially under the constraints of our relatively small physics dataset, we design adaptive mechanisms that ensure continuous learnability throughout the training process. Specifically, we propose two complementary strategies: \emph{preliminary pass rate filtering} and \emph{adaptive exploration space expansion}.

\paragraph{Preliminary Pass Rate Filtering.}  
Data quality directly impacts learnability, and even after extensive curation, physics datasets may still contain problematic samples, such as tasks relying on unavailable diagrams or incomplete context. To mitigate this issue, we apply a filtering procedure before training, based on pass rate statistics. Concretely, we perform rollouts on the training dataset using the Qwen3-30B-A3B-Thinking model under a pass@88 setting, and exclude tasks that are either too easy (pass rate $>0.7$) or too difficult (pass rate $=0$). Formally, the retained dataset is:
\begin{equation}
    \mathcal{D}_{\text{filtered}} = \{\, q \in \mathcal{D} \;|\; 0 < \text{pass}(q) \leq 0.7 \,\}.
\end{equation}

This filtering serves two purposes:  
\begin{enumerate}[left=0pt]
    \item \textbf{Removing tasks with $\text{pass}=0$ or $\text{pass}=1$ prevents zero-learnability cases.} As defined in GSPO~\citep{zheng2025gspo}, when all group samples share the same outcome, the estimated advantage ($\hat{A}^{\text{GSPO}}$) becomes zero, eliminating effective learning signals. Filtering these tasks mitigates reward sparsity.
    \item \textbf{Removing overly easy tasks (pass $>0.7$) prevents entropy collapse.} According to the entropy mechanism analysis~\citep{cui2025entropy}, when the majority of tasks can be solved with high confidence, RL updates push the policy toward low-entropy solutions, accelerating collapse. Given that modern base models already possess strong reasoning capabilities, excluding trivial tasks helps maintain diversity and prevents premature convergence.
\end{enumerate}

% After pass@88 filtering, our dataset is reduced to approximately 2,000 tasks that balance non-trivial difficulty with solvability, thereby ensuring data-level learnability in subsequent training.

\paragraph{Adaptive Exploration Space Expansion.}  
Even with curated data, performance bottlenecks can also arise from insufficient exploration~\citep{cui2025entropy,deepscaler2025}. As the model improves, fixed rollout configurations may fail to provide adequate opportunities for exploration, leading to stagnation. To address this, we dynamically expand the exploration space in line with the model’s evolving capability, thereby sustaining learnability.

We consider two complementary dimensions of expansion:

\begin{enumerate}[left=0pt]
    \item \textbf{Group size expansion.}  
    In the GSPO~\citep{zheng2025gspo} group-based advantage estimation framework, each training question $q$ is evaluated with a group of $G$ sampled responses $\{o_i\}_{i=1}^G \sim \pi_{\theta_{\text{old}}}(\cdot|q)$, and the relative rewards within the group are used to construct standardized advantages. Increasing the group size $G$ enhances the probability of generating at least one high-quality trajectory, especially for difficult problems where success rates are intrinsically low. Larger $G$ thus provides a stronger learning signal by ensuring that rare but informative trajectories are more likely to appear within each group, which in turn amplifies effective gradient updates during training. This mechanism directly alleviates reward sparsity and strengthens the stability of group-based advantage estimation.

    \item \textbf{Generation window expansion.}  
    The maximum output length (generation window) limits the depth of reasoning that can be expressed. If too short, the model’s reasoning chains for complex physics problems are truncated, producing incomplete or incorrect answers. We gradually extend the generation window as training progresses, allowing the model to explore longer and more coherent reasoning chains. This expansion improves solvability for high-complexity problems and reduces truncation-induced errors.
\end{enumerate}

% Notably, $n$-sample expansion and generation window expansion act in a complementary manner: the former enhances trajectory diversity, while the latter guarantees trajectory completeness. Together, they provide the model with a broader and deeper exploration space, ensuring sustained learnability and preventing stagnation.

% \paragraph{Summary.}  
% By combining preliminary pass rate filtering with dynamic exploration space expansion, we maintain non-trivial but solvable training signals and adapt exploration opportunities to the model’s capability. These mechanisms jointly enable stable and continuous performance improvement on physics Olympiad-level tasks.

\subsubsection{Training Stabilization Mechanism}

\paragraph{Mitigate Train-inference Mismatch}
Recent studies~\citep{yao2025tis,jiacai2025speed} have noticed that the train-inference engine difference is a key cause to instability in training. To formally understand this statement, we rewrite Eq.~\ref{eq:pg_loss} as,

\begin{equation}
    \nabla_\theta J(\pi_\theta) = \mathbb{E}_{\tau \sim \pi_\theta^{rollout}}\Bigg[\sum_{t=0}^{T} \nabla_\theta \log \pi_\theta^{train}(a_t \mid s_t) \, A^\pi(s_t, a_t)\Bigg],
\end{equation}

where $\pi_\theta^{rollout}$ denote the policy used to generate trajectories during rollout, and $\pi_\theta^{train}$ denote the policy evaluated during gradient computation. Modern RL frameworks~\citep{slime_github,sheng2024verl} often adopt different engines for rollout (e.g. vllm~\citep{kwon2023vllm} and SGLang~\citep{zheng2024sglang}) and 
training (e.g. FSDP~\citep{merry2021fsdp} and Megatron~\citep{shoeybi2019megatron}). While this design significantly improve throughout, it  inadvertently introduce mismatch between $\pi_\theta^{rollout}$ and $\pi_\theta^{train}$ due to differences in numerical precision, computation optimization strategies, and
kernel implementations, leading to biased gradient estimates and training instability. 

\begin{equation}
    \pi_\theta^{rollout}(a|s) \neq \pi_\theta^{train}(a|s)
\end{equation}

To mitigate this mismatch and stablize training, we adopt Truncated Importance Sampling (TIS)~\citep{yao2025tis}, which applies importance weighting to rebalance gradients computed under the training policy $\pi_\theta^{train}$ using trajectories sampled from the rollout policy $\pi_\theta^{rollout}$.

\begin{equation}
    \nabla_{\theta}J(\pi_{\theta}) = \mathbb{E}_{\tau \sim \pi_\theta^{rollout}} \left[ \sum_{t=0}^{T} \min\left( \frac{\pi_\theta^{train}(a_t|s_t)}{\pi_\theta^{rollout}(a_t|s_t)}, C \right) \nabla_{\theta} \log \pi_{\theta}(a_t|s_t) A^{\pi}(s_t, a_t) \right],
\end{equation}

where $C$ is a truncation hyperparameter that controls the variance of the importance weights. The truncation operator $\min(\cdot, C)$ prevents excessively large weights that could destabilize training, while still correcting for the distributional shift.

\subsection{Training Dynamics}\label{train_dyn}

\paragraph{Implementation.} For implementation of P1 training pipeline, we adopt the slime~\citep{slime_github} framework, which is a efficient LLMs post-training framework connecting Megatron with SGLang. As for base model, we use \texttt{Qwen3-30B-A3B-Thinking} and \texttt{Qwen3-235B-A22B-Thinking} as starting points of our \texttt{P1-30B-A3B} and \texttt{P1-235B-A22B}. To adaptively adjust  aforementioned learnability during training, we periodically resume training from previous checkpoint with updated configurations. 

\paragraph{Settings of Different Phrases.}
We report the settings of different stages during the post training process of P1 models. As shown in the table, we gradually expand the exploration space by increasing group size and generation window. We also apply the train-inference correction technique during the whole multi-stage process. Notely, we build up our algorithm upon GSPO to stabilize the training of MoE models. Besides, we only enable the rule-based verifier during training, the reason for which will be discussed in Section~\ref{sec:xverify}.

\begin{table}[thbp]
\centering
\caption{Configuration of different phrases in P1 training.}
\label{tab:training_params}
\resizebox{0.8\linewidth}{!}{
\begin{tabular}{lccc|cccc}
\toprule
\multirow{2}{*}{\textbf{Model}} & \multicolumn{3}{c|}{\textbf{P1-30B-A3B}} & \multicolumn{4}{c}{\textbf{P1-235B-A22B}} \\
\cmidrule(lr){2-4} \cmidrule(lr){5-8}
& Stage 1 & Stage 2 & Stage 3 & Stage 1 & Stage 2 & Stage 3 & Stage 4 \\
\midrule
\textbf{Group size} & 16 & 32 & 32 & 16 & 32 & 32 & 32 \\
\textbf{Generation Window} & 48k & 48k & 64k & 48k & 48k & 64k & 80k \\
\textbf{Lr} & \multicolumn{3}{c|}{$1 \times 10^{-6}$} & \multicolumn{4}{c}{$1 \times 10^{-6}$} \\
\textbf{Algorithm} & \multicolumn{3}{c|}{GSPO w. tis} & \multicolumn{4}{c}{GSPO w. tis} \\
\textbf{Verifier Choice} & \multicolumn{3}{c|}{Rule-based Verifier Only} & \multicolumn{4}{c}{Rule-based Verifier Only} \\
\textbf{Rollout Batch Size} & \multicolumn{3}{c|}{2048} & \multicolumn{4}{c}{1024} \\
\textbf{Update Batch Size} & \multicolumn{3}{c|}{512} & \multicolumn{4}{c}{256} \\
\bottomrule
\end{tabular}
}
\end{table}

\begin{figure}[t]
    \centering
    \subfloat{\includegraphics[width=0.48\textwidth]{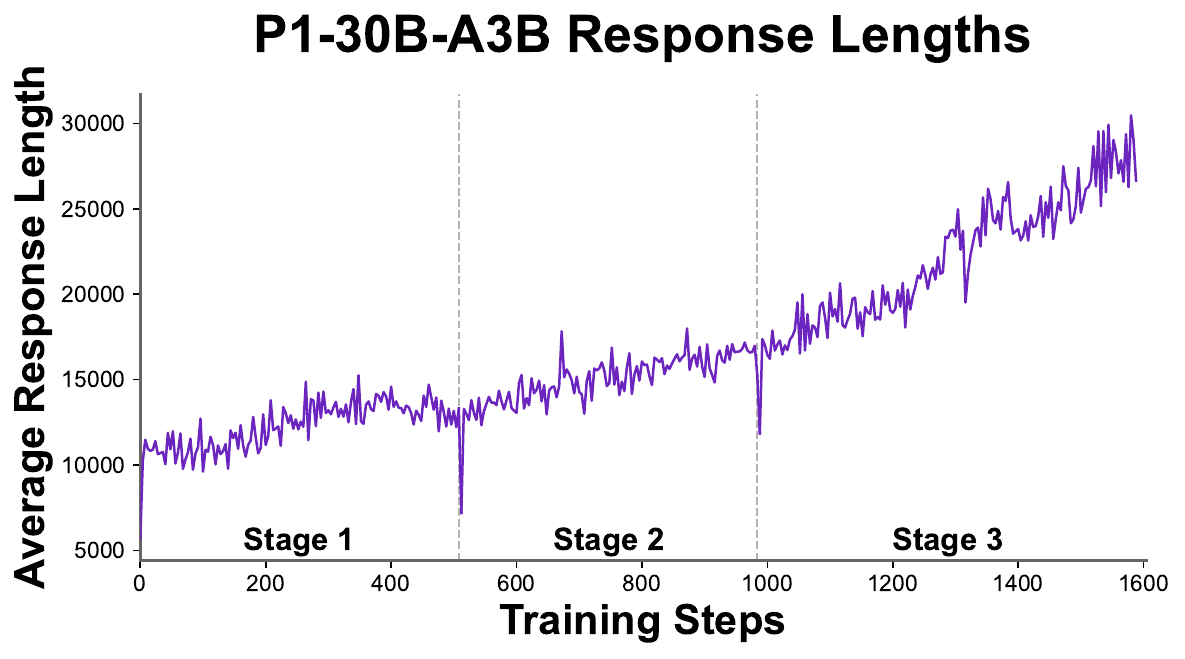}\label{fig:image1}}
    \hfill
    \subfloat{\includegraphics[width=0.48\textwidth]{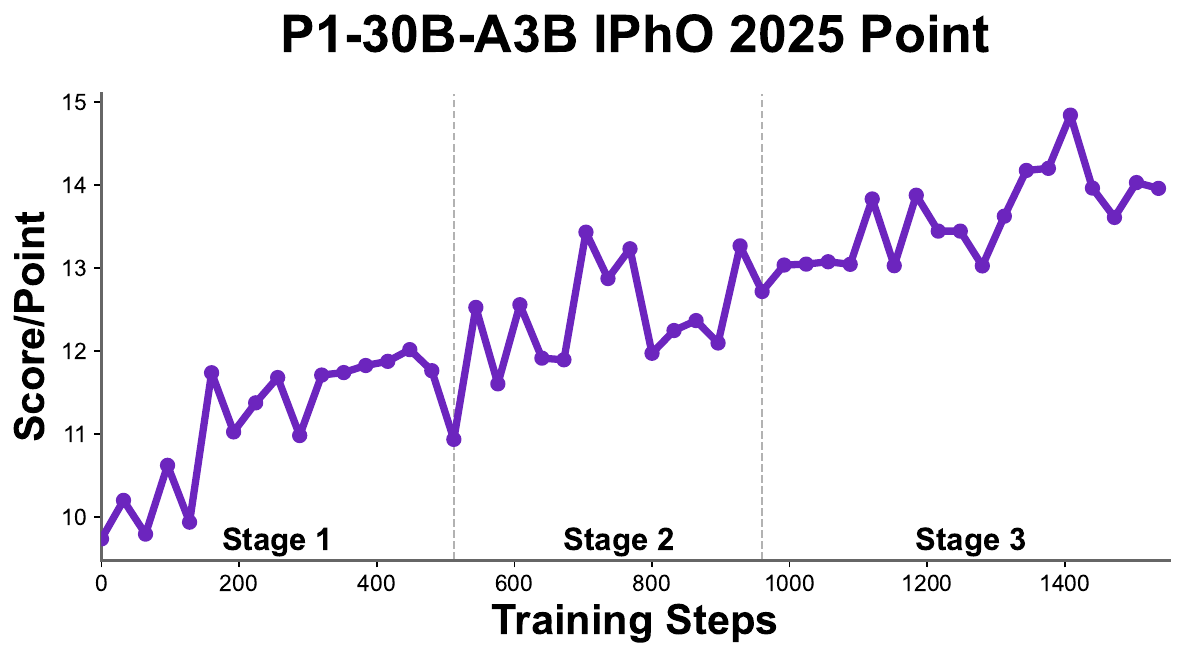}\label{fig:image2}}
    
    % \vspace{0.5cm}
    
    \subfloat{\includegraphics[width=0.48\textwidth]{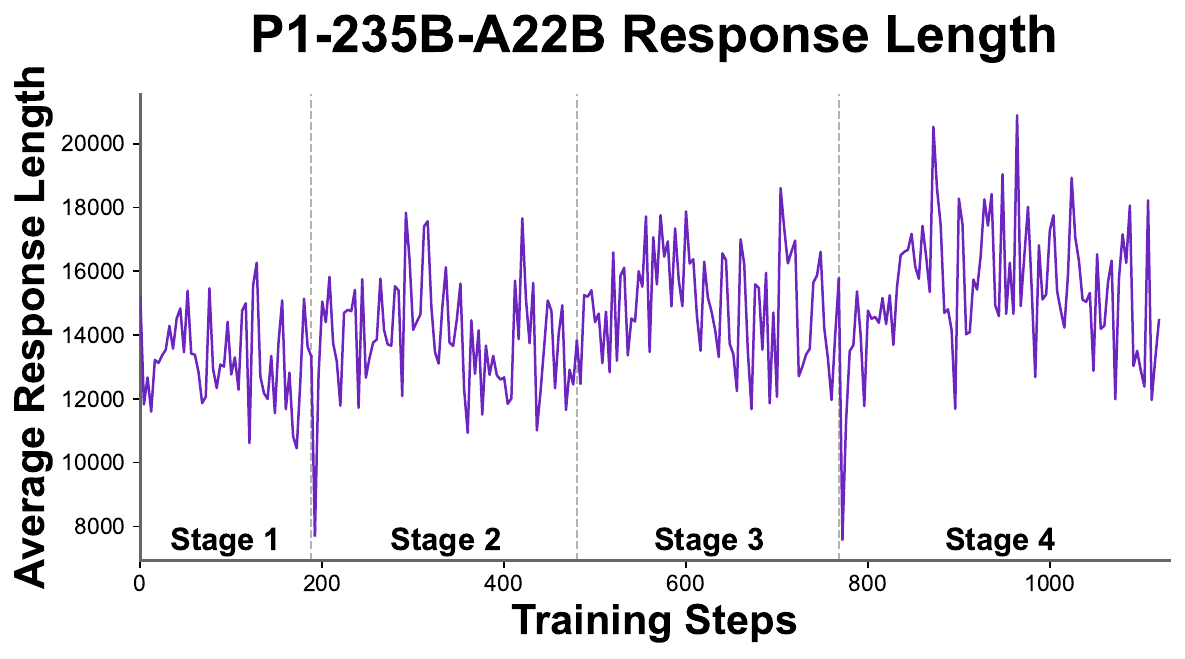}\label{fig:image3}}
    \hfill
    \subfloat{\includegraphics[width=0.48\textwidth]{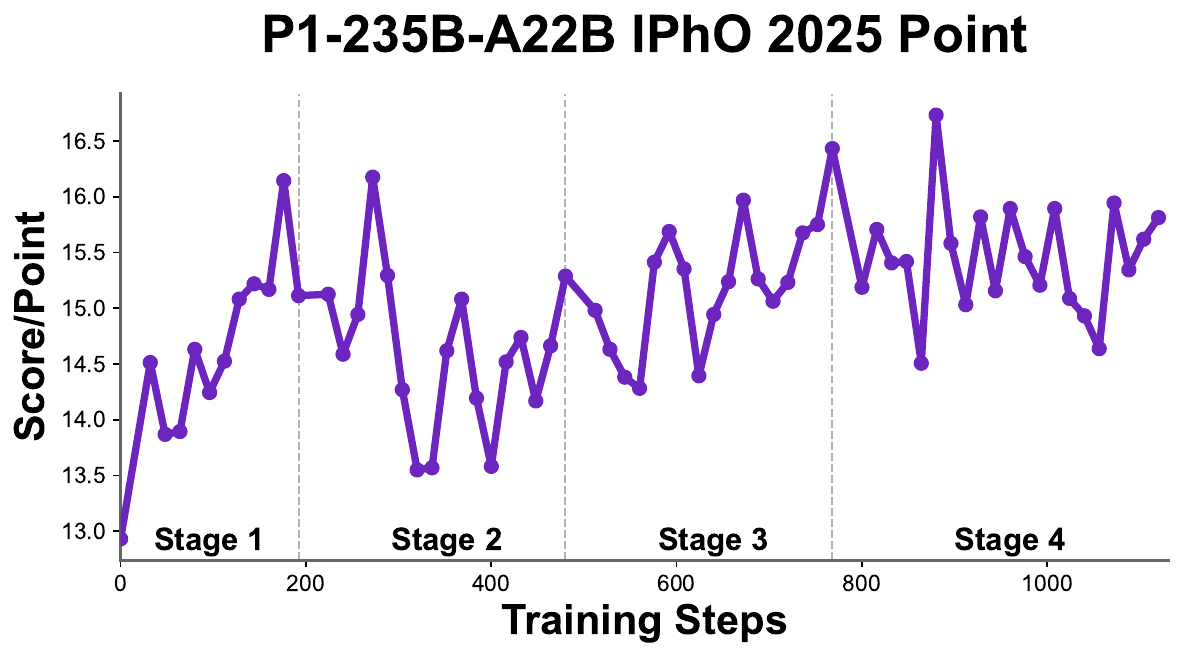} \label{fig:image4}}
    
    \caption{Training dynamics of P1 models. The upper side shows dynamics of P1-30B and the lower side shows the P1-235B variant. The LHS present the average response length on train dataset during RL training process. The RHS present the evaluation results on IPhO 2025 on model checkpoints during training, please note that evaluation here leverage rule-based verifier and model-based verifier (Qwen3-30B-Instruct-2507). }
    \label{fig:training_dynamic}
\end{figure}

\paragraph{Analysis.}
As shown in Figure~\ref{fig:training_dynamic}, we present training dynamics of multi-stages RL training. It can be observed that the response length of P1 gradually increases during training, revealing the model's ability to reason and solve complex problems getting more deeply and requires a larger exploration space to explore. Therefore, we gradually expand the exploration space by increasing generation window and group size by switching to different stages. Furthermore, as for validation results, we can observe that the P1 models experience an steady improvement on IPhO 2025 during training, revealing the effectiveness and stability of our training algorithm.

\subsection{Agentic Augmentation}

During inference, we scale up test-time effort by incorporating multi-agent framework. We directly apply P1 model into \texttt{PhysicsMinions}~\citep{physicsminions}, which is a latest agentic framework designed for complex physics reasoning. \texttt{PhysicsMinions} consists of three coevolutionary studios: the \emph{Visual Studio}, the \emph{Logic Studio}, and the \emph{Review Studio}. Given a multimodal problem with diagrams or plots, the Visual Studio first observes, validates, and reflects on the input to extract structured information, which is then passed to the Logic Studio. In the Logic Studio, a solver generates an initial solution and an introspector refines it through self-improvement before passing it on. The Review Studio then applies dual-stage verification: the Physics-Verifier checks physical consistency (e.g., constants and units), while the General-Verifier conducts more detailed inspections of logic, reasoning, and calculations. 

If either verification stage fails, a detailed bug report is returned to the Logic Studio, where the introspector revises the solution and resubmits it to Review Studio for verification. This process repeats until the solution passes a predefined number of consecutive verifications (CV), which is the only hyperparameter in the system. A solution that passes CV checks consecutively is accepted as the final solution; if it fails CV times consecutively, the solver regenerates a new candidate solution. This collaborative critique-and-refine cycle defines the system's coevolutionary process, with CV set to 2 by default. It's noteworthy that because P1 models are text-only LLMs, we disable the Visual Studio for adaptivity. In simple words, we instantiate the solver in the Logic Studio as well as the dual verifiers in the Review Studio using P1 models within \texttt{PhysicsMinions}.
\begin{table}[th]
\centering
\vspace{-10pt}
\caption{Evaluation results on the HiPhO benchmark (13 physics Olympiads from 2024--2025) using the \textit{exam score} metric. \medalbox{Gold!50}{Gold}, \medalbox{Silver!70}{Silver} and \medalbox{Bronze!40}{Bronze} indicate scores above the respective thresholds. Models are ranked by medal counts; \textbf{bold} is the highest score, and \underline{underline} is the second highest. Here, only the theoretical parts of exams are used, hence Full Mark (Model) $\leq$ Full Mark (Human).}
\vspace{-10pt}
\label{tab:hipho_all}
\small
\resizebox{\textwidth}{!}{%
\setlength{\tabcolsep}{2.3pt}

\begin{tabular}{lccccccccccccc|c|ccc}
\toprule
\textbf{Physics Olympiad} & \multicolumn{2}{c}{\textbf{IPhO}} & \textbf{APhO} & \multicolumn{2}{c}{\textbf{EuPhO}} & \multicolumn{2}{c}{\textbf{NBPhO}} & \multicolumn{6}{c|}{\textbf{PanPhO}~~
\textbf{\footnotesize PanMechanics} ~~ \textbf{F=MA}} & \textbf{Avg.} & \multicolumn{3}{c}{\textbf{Medal}} \\
\textbf{Year} &2025&2024&2025&2025&2024&2025&2024&2025&2024&2025&2024&2025&2024& & \multicolumn{3}{c}{\textbf{Table}} \\ \midrule
Full Mark (Human) & 30.0 & 30.0 & 30.0 & 30.0 & 30.0 & 72.0 & 72.0 & 100.0 & 100.0 & 100.0 & 100.0 & 25.0 & 25.0 & 57.2 \\
Full Mark (Model) & 29.4 & 29.3 & 30.0 & 29.0 & 28.0 & 43.5 & 50.0 & 100.0 & 98.0 & 100.0 & 100.0 & 25.0 & 25.0 & 52.9 \\
Top-1 Score (Human) & 29.2 & 29.4 & 30.0 & 27.0 & 30.0 & 53.2 & 40.8 & 81.0 & 66.5 & 62.0 & 51.0 & 25.0 & 24.0 & 42.2 \\
Top-1 Score (Compared Models) & 22.7 & 25.9 & 27.9 & 14.9 & 23.4 & 34.1 & 35.9 & 60.3 & 75.4 & 72.1 & 79.0 & 22.8 & 22.4 & 39.8 \\
\rowcolor{Gold!40}
Gold Medal   & 19.7 & 20.8 & 23.3 & 16.5 & 20.4 & 28.6 & 26.5 & 41.5 & 52.0 & 52.0 & 51.0 & 15.0 & 14.0 & \cellcolor{white}29.3 & \multicolumn{3}{>{\cellcolor{white}}l}{\textcolor{Gold}{\faMedal}}  \\
\rowcolor{Silver!60}
Silver Medal & 12.1 & 11.1 & 18.7 & 9.8 & 14.2 & 20.1 & 19.4 & 28.5 & 37.5 & 36.0 & 26.0 & 11.0 & 12.0 & \cellcolor{white}19.7 & \multicolumn{3}{>{\cellcolor{white}}c}{\textcolor{Silver}{\faMedal}}  \\
\rowcolor{Bronze!40}
Bronze Medal & 7.2  & 3.6  & 13.1 & 5.8  & 8.9  & 15.2  & 13.5  & 14.5 & 16.0 & 20.0 & 12.0 & 9.0 & 10.0 & \cellcolor{white}11.4 & \multicolumn{3}{>{\cellcolor{white}}r}{\textcolor{Bronze}{\faMedal}}  \\ \midrule
\textcolor[HTML]{6c25be}{\textbf{\large P1-235B-A22B + PhysicsMinions}} &\cellcolor{Gold!35}\textbf{23.2}&\cellcolor{Gold!35}\underline{25.2}&\cellcolor{Gold!35}\textbf{28.0}&\cellcolor{Silver!60}12.4&\cellcolor{Gold!35}\textbf{23.5}&\cellcolor{Gold!35}31.9&\cellcolor{Gold!35}\underline{35.4}&\cellcolor{Gold!35}\underline{57.7}&\cellcolor{Gold!35}67.0&\cellcolor{Gold!35}\textbf{77.5}&\cellcolor{Gold!35}74.8&\cellcolor{Gold!35}21.5&\cellcolor{Gold!35}20.5&\textbf{38.4}& ~\mc{Gold}{12} & \mc{Silver}{1} & \mc{Bronze}{0}\\
Gemini-2.5-Pro &\cellcolor{Gold!35}\underline{22.7}&\cellcolor{Gold!35}\textbf{25.9}&\cellcolor{Gold!35}\underline{27.9}&\cellcolor{Silver!60}\textbf{14.9}&\cellcolor{Gold!35}21.8&\cellcolor{Gold!35}32.3&\cellcolor{Gold!35}\textbf{35.9}&\cellcolor{Gold!35}\textbf{60.3}&\cellcolor{Gold!35}64.1&\cellcolor{Gold!35}69.5&\cellcolor{Gold!35}70.2&\cellcolor{Gold!35}\textbf{22.8}&\cellcolor{Gold!35}\underline{22.0}&\underline{37.7}& ~\mc{Gold}{12} & \mc{Silver}{1} & \mc{Bronze}{0}\\
\textcolor[HTML]{6c25be}{\textbf{\large P1-235B-A22B}} &\cellcolor{Gold!35}21.2&\cellcolor{Gold!35}24.7&\cellcolor{Gold!35}27.4&\cellcolor{Silver!60}10.8&\cellcolor{Gold!35}23.0&\cellcolor{Gold!35}31.8&\cellcolor{Gold!35}28.4&\cellcolor{Gold!35}54.7&\cellcolor{Gold!35}56.7&\cellcolor{Gold!35}\underline{74.7}&\cellcolor{Gold!35}72.9&\cellcolor{Gold!35}20.9&\cellcolor{Gold!35}19.4&35.9& ~\mc{Gold}{12} & \mc{Silver}{1} & \mc{Bronze}{0}\\
Gemini-2.5-Flash-Thinking &\cellcolor{Gold!35}20.2&\cellcolor{Gold!35}23.9&\cellcolor{Gold!35}27.4&\cellcolor{Silver!60}\underline{13.2}&\cellcolor{Gold!35}21.9&\cellcolor{Gold!35}29.0&\cellcolor{Gold!35}29.3&\cellcolor{Gold!35}44.6&\cellcolor{Gold!35}54.9&\cellcolor{Gold!35}60.5&\cellcolor{Gold!35}55.9&\cellcolor{Gold!35}17.8&\cellcolor{Gold!35}19.1&32.1& ~\mc{Gold}{12} & \mc{Silver}{1} & \mc{Bronze}{0}\\
GPT-5 &\cellcolor{Gold!35}22.3&\cellcolor{Silver!60}20.2&\cellcolor{Gold!35}27.0&\cellcolor{Silver!60}10.3&\cellcolor{Gold!35}21.7&\cellcolor{Gold!35}32.9&\cellcolor{Gold!35}32.8&\cellcolor{Gold!35}55.9&\cellcolor{Gold!35}\underline{69.8}&\cellcolor{Gold!35}69.4&\cellcolor{Gold!35}\textbf{79.0}&\cellcolor{Gold!35}\underline{22.4}&\cellcolor{Gold!35}\textbf{22.4}&37.4& ~\mc{Gold}{11} & \mc{Silver}{2} & \mc{Bronze}{0} \\
%
% Qwen3-VL-235B-A22B-Thinking &\cellcolor{Gold!35}21.2&\cellcolor{Silver!60}19.8&\cellcolor{Gold!35}27.2&\cellcolor{Silver!60}11.9&\cellcolor{Gold!35}\underline{23.4}&\cellcolor{Gold!35}33.3&\cellcolor{Gold!35}31.7&\cellcolor{Gold!35}56.7&\cellcolor{Gold!35}59.9&\cellcolor{Gold!35}70.4&\cellcolor{Gold!35}71.1&\cellcolor{Gold!35}21.5&\cellcolor{Gold!35}21.2&36.1& ~\mc{Gold}{11} & \mc{Silver}{2} & \mc{Bronze}{0} \\
%
o3 &\cellcolor{Silver!60}15.7&\cellcolor{Gold!35}23.7&\cellcolor{Gold!35}25.9&\cellcolor{Silver!60}11.4&\cellcolor{Gold!35}21.6&\cellcolor{Gold!35}\textbf{34.1}&\cellcolor{Gold!35}33.5&\cellcolor{Gold!35}47.3&\cellcolor{Gold!35}55.9&\cellcolor{Gold!35}71.4&\cellcolor{Gold!35}75.6&\cellcolor{Gold!35}22.0&\cellcolor{Gold!35}20.6&35.3& ~\mc{Gold}{11} & \mc{Silver}{2} & \mc{Bronze}{0}\\
Grok-4 &\cellcolor{Silver!60}18.7&\cellcolor{Gold!35}23.5&\cellcolor{Gold!35}25.0&\cellcolor{Silver!60}11.5&\cellcolor{Gold!35}20.5&\cellcolor{Silver!60}25.8&\cellcolor{Gold!35}29.3&\cellcolor{Gold!35}45.0&\cellcolor{Gold!35}\textbf{75.4}&\cellcolor{Gold!35}72.1&\cellcolor{Gold!35}\underline{78.6}&\cellcolor{Gold!35}19.8&\cellcolor{Gold!35}19.8&35.8& ~\mc{Gold}{10} & \mc{Silver}{3} & \mc{Bronze}{0}\\
Qwen3-235B-A22B-Thinking-2507 &\cellcolor{Silver!60}17.1&\cellcolor{Gold!35}23.0&\cellcolor{Gold!35}26.2&\cellcolor{Silver!60}10.9&\cellcolor{Gold!35}20.4&\cellcolor{Gold!35}\underline{33.6}&\cellcolor{Gold!35}28.1&\cellcolor{Gold!35}44.7&\cellcolor{Silver!60}51.8&\cellcolor{Gold!35}69.1&\cellcolor{Gold!35}72.9&\cellcolor{Gold!35}18.5&\cellcolor{Gold!35}18.9&33.5& ~\mc{Gold}{10} & \mc{Silver}{3} & \mc{Bronze}{0}\\
DeepSeek-R1 &\cellcolor{Silver!60}18.5&\cellcolor{Gold!35}24.6&\cellcolor{Gold!35}25.4&\cellcolor{Silver!60}10.8&\cellcolor{Gold!35}21.4&\cellcolor{Silver!60}26.3&\cellcolor{Silver!60}20.5&\cellcolor{Gold!35}42.2&\cellcolor{Silver!60}47.4&\cellcolor{Gold!35}65.4&\cellcolor{Gold!35}72.5&\cellcolor{Gold!35}18.3&\cellcolor{Gold!35}18.5&31.7& ~\mc{Gold}{8} & \mc{Silver}{5} & \mc{Bronze}{0} \\
Claude-4-Sonnet-Thinking &\cellcolor{Silver!60}19.0&\cellcolor{Gold!35}22.0&\cellcolor{Gold!35}24.8&\cellcolor{Bronze!40}9.7&\cellcolor{Gold!35}20.5&\cellcolor{Silver!60}28.1&\cellcolor{Silver!60}25.6&\cellcolor{Gold!35}43.1&\cellcolor{Silver!60}39.3&\cellcolor{Gold!35}57.4&\cellcolor{Gold!35}61.8&\cellcolor{Gold!35}19.2&\cellcolor{Gold!35}20.1&30.0& ~\mc{Gold}{8} & \mc{Silver}{4} & \mc{Bronze}{1}\\
\textcolor[HTML]{6c25be}{\textbf{\large P1-30B-A3B}} &\cellcolor{Silver!60}18.5&\cellcolor{Gold!35}22.3&\cellcolor{Gold!35}25.4&\cellcolor{Bronze!40}7.4&\cellcolor{Silver!60}18.8&\cellcolor{Gold!35}29.2&\cellcolor{Silver!60}24.0&\cellcolor{Gold!35}47.0&\cellcolor{Silver!60}51.4&\cellcolor{Gold!35}69.5&\cellcolor{Gold!35}69.1&\cellcolor{Gold!35}19.6&\cellcolor{Gold!35}20.0&32.5& ~\mc{Gold}{8} & \mc{Silver}{4} & \mc{Bronze}{1}\\
Qwen3-235B-A22B &\cellcolor{Silver!60}17.8&\cellcolor{Gold!35}23.8&\cellcolor{Gold!35}26.0&\cellcolor{Bronze!40}9.2&\cellcolor{Gold!35}21.5&\cellcolor{Silver!60}28.4&\cellcolor{Gold!35}31.1&\cellcolor{Gold!35}42.3&\cellcolor{Silver!60}49.1&\cellcolor{Gold!35}63.1&\cellcolor{Silver!60}44.6&\cellcolor{Gold!35}18.4&\cellcolor{Gold!35}17.4&30.2& ~\mc{Gold}{8} & \mc{Silver}{4} & \mc{Bronze}{1} \\
Qwen3-32B &\cellcolor{Silver!60}15.7&\cellcolor{Silver!60}19.3&\cellcolor{Gold!35}23.9&\cellcolor{Silver!60}9.8&\cellcolor{Gold!35}21.2&\cellcolor{Gold!35}28.9&\cellcolor{Silver!60}24.1&\cellcolor{Silver!60}36.6&\cellcolor{Silver!60}41.8&\cellcolor{Gold!35}67.0&\cellcolor{Gold!35}59.2&\cellcolor{Gold!35}18.9&\cellcolor{Gold!35}16.6&29.5& ~\mc{Gold}{7} & \mc{Silver}{6} & \mc{Bronze}{0} \\
o4-mini &\cellcolor{Silver!60}15.4&\cellcolor{Gold!35}22.9&\cellcolor{Silver!60}22.8&\cellcolor{Silver!60}10.1&\cellcolor{Gold!35}20.9&\cellcolor{Silver!60}26.9&\cellcolor{Gold!35}27.3&\cellcolor{Silver!60}39.4&\cellcolor{Silver!60}47.1&\cellcolor{Gold!35}64.2&\cellcolor{Gold!35}62.5&\cellcolor{Gold!35}18.6&\cellcolor{Gold!35}18.5&30.5& ~\mc{Gold}{7} & \mc{Silver}{6} & \mc{Bronze}{0} \\
o4-mini (high) &\cellcolor{Silver!60}16.0&\cellcolor{Gold!35}23.7&\cellcolor{Silver!60}22.9&\cellcolor{Silver!60}12.0&\cellcolor{Silver!60}20.1&\cellcolor{Silver!60}27.4&\cellcolor{Gold!35}29.8&\cellcolor{Silver!60}41.4&\cellcolor{Silver!60}50.9&\cellcolor{Gold!35}69.1&\cellcolor{Gold!35}67.3&\cellcolor{Gold!35}18.6&\cellcolor{Gold!35}18.8&32.2& ~\mc{Gold}{6} & \mc{Silver}{7} & \mc{Bronze}{0}\\
Qwen3-30B-A3B-Thinking-2507 &\cellcolor{Silver!60}15.6&\cellcolor{Silver!60}19.7&\cellcolor{Gold!35}23.5&\cellcolor{Bronze!40}7.4&\cellcolor{Silver!60}16.4&\cellcolor{Gold!35}28.6&\cellcolor{Silver!60}22.2&\cellcolor{Silver!60}40.5&\cellcolor{Silver!60}43.8&\cellcolor{Gold!35}67.7&\cellcolor{Gold!35}66.5&\cellcolor{Gold!35}18.3&\cellcolor{Gold!35}18.0&29.9& ~\mc{Gold}{6} & \mc{Silver}{6} & \mc{Bronze}{1}\\
Kimi-K2-Thinking* &\cellcolor{Silver!60}19.1&\cellcolor{Gold!35}22.8&\cellcolor{Silver!60}22.4&5.2&\cellcolor{Gold!35}20.9&\cellcolor{Silver!60}26.0&\cellcolor{Silver!60}20.6&\cellcolor{Gold!35}52.5&\cellcolor{Silver!60}51.3&\cellcolor{Silver!60}47.3&\cellcolor{Gold!35}64.8&\cellcolor{Gold!35}20.6&\cellcolor{Gold!35}19.7&30.2& ~\mc{Gold}{6} & \mc{Silver}{6} & \mc{Bronze}{0} \\
Kimi-K2-Instruct &\cellcolor{Silver!60}16.5&\cellcolor{Silver!60}19.8&\cellcolor{Gold!35}24.2&\cellcolor{Silver!60}11.0&\cellcolor{Silver!60}16.9&\cellcolor{Silver!60}26.5&\cellcolor{Silver!60}26.2&\cellcolor{Silver!60}35.9&\cellcolor{Silver!60}41.8&\cellcolor{Gold!35}65.9&\cellcolor{Gold!35}58.9&\cellcolor{Gold!35}16.0&\cellcolor{Gold!35}18.2&29.1& ~\mc{Gold}{5} & \mc{Silver}{8} & \mc{Bronze}{0} \\
GPT-OSS-120B &\cellcolor{Silver!60}16.9&\cellcolor{Gold!35}21.4&\cellcolor{Silver!60}22.8&\cellcolor{Bronze!40}9.1&\cellcolor{Silver!60}19.9&\cellcolor{Silver!60}26.0&\cellcolor{Silver!60}25.8&\cellcolor{Silver!60}37.4&\cellcolor{Silver!60}41.8&\cellcolor{Gold!35}57.1&\cellcolor{Gold!35}59.7&\cellcolor{Gold!35}17.8&\cellcolor{Gold!35}17.6&28.7& ~\mc{Gold}{5} & \mc{Silver}{7} & \mc{Bronze}{1} \\
Intern-S1 &\cellcolor{Silver!60}15.9&\cellcolor{Silver!60}14.2&\cellcolor{Silver!60}21.7&\cellcolor{Bronze!40}9.0&\cellcolor{Silver!60}16.6&\cellcolor{Silver!60}23.0&\cellcolor{Silver!60}20.5&\cellcolor{Silver!60}41.1&\cellcolor{Silver!60}50.3&\cellcolor{Gold!35}60.4&\cellcolor{Gold!35}57.4&\cellcolor{Gold!35}18.4&\cellcolor{Gold!35}19.5&28.3& ~\mc{Gold}{4} & \mc{Silver}{8} & \mc{Bronze}{1} \\
Qwen3-30B-A3B &\cellcolor{Silver!60}13.6&\cellcolor{Silver!60}15.4&\cellcolor{Silver!60}22.7&\cellcolor{Silver!60}9.8&\cellcolor{Silver!60}16.5&\cellcolor{Silver!60}24.7&\cellcolor{Silver!60}21.7&\cellcolor{Silver!60}31.9&\cellcolor{Silver!60}39.5&\cellcolor{Silver!60}49.9&\cellcolor{Silver!60}45.0&\cellcolor{Gold!35}15.5&\cellcolor{Gold!35}15.0&24.7& ~\mc{Gold}{2} & \mc{Silver}{11} & \mc{Bronze}{0} \\
Claude-4-Sonnet &\cellcolor{Silver!60}15.7&\cellcolor{Silver!60}19.2&\cellcolor{Silver!60}22.8&\cellcolor{Bronze!40}9.5&\cellcolor{Silver!60}16.5&\cellcolor{Silver!60}27.5&\cellcolor{Silver!60}21.3&\cellcolor{Silver!60}40.4&\cellcolor{Silver!60}43.3&\cellcolor{Silver!60}46.5&\cellcolor{Silver!60}48.5&\cellcolor{Gold!35}16.8&\cellcolor{Gold!35}16.5&26.5& ~\mc{Gold}{2} & \mc{Silver}{10} & \mc{Bronze}{1}\\
DeepSeek-V3 &\cellcolor{Silver!60}13.6&\cellcolor{Silver!60}16.4&\cellcolor{Silver!60}22.1&\cellcolor{Bronze!40}7.1&\cellcolor{Silver!60}17.2&\cellcolor{Silver!60}21.1&\cellcolor{Bronze!40}17.3&\cellcolor{Silver!60}37.2&\cellcolor{Bronze!40}35.0&\cellcolor{Silver!60}48.4&\cellcolor{Silver!60}46.5&\cellcolor{Silver!60}14.1&\cellcolor{Gold!35}15.6&24.0& ~\mc{Gold}{1} & \mc{Silver}{9} & \mc{Bronze}{3} \\
Mistral-Medium-3 &\cellcolor{Silver!60}14.2&\cellcolor{Silver!60}14.1&\cellcolor{Silver!60}19.9&\cellcolor{Bronze!40}8.5&\cellcolor{Bronze!40}12.2&\cellcolor{Silver!60}20.4&\cellcolor{Silver!60}19.6&\cellcolor{Silver!60}30.8&\cellcolor{Bronze!40}28.6&\cellcolor{Bronze!40}32.9&\cellcolor{Silver!60}36.1&\cellcolor{Silver!60}13.9&\cellcolor{Gold!35}14.1&20.4& ~\mc{Gold}{1} & \mc{Silver}{8} & \mc{Bronze}{4}\\
GPT-4o &\cellcolor{Bronze!40}10.2&\cellcolor{Bronze!40}9.4&\cellcolor{Bronze!40}15.1&\cellcolor{Bronze!40}6.8&\cellcolor{Bronze!40}9.2&\cellcolor{Bronze!40}16.4&11.7&\cellcolor{Bronze!40}27.8&\cellcolor{Bronze!40}22.8&\cellcolor{Bronze!40}28.2&\cellcolor{Silver!60}26.5&\cellcolor{Gold!35}15.0&\cellcolor{Bronze!40}10.9&16.2& ~\mc{Gold}{1} & \mc{Silver}{1} & \mc{Bronze}{10}\\
InternVL3-78B-Instruct &\cellcolor{Silver!60}12.9&\cellcolor{Silver!60}12.5&\cellcolor{Bronze!40}17.7&\cellcolor{Bronze!40}7.5&\cellcolor{Silver!60}15.2&\cellcolor{Silver!60}22.3&\cellcolor{Silver!60}22.5&\cellcolor{Bronze!40}26.2&\cellcolor{Bronze!40}27.4&\cellcolor{Bronze!40}21.1&\cellcolor{Silver!60}27.1&\cellcolor{Silver!60}12.0&\cellcolor{Silver!60}13.0&18.3& ~\mc{Gold}{0} & \mc{Silver}{8} & \mc{Bronze}{5}\\
GLM-4.5V &\cellcolor{Bronze!40}11.9&\cellcolor{Bronze!40}4.4&\cellcolor{Bronze!40}16.2&\cellcolor{Bronze!40}8.7&\cellcolor{Bronze!40}14.1&\cellcolor{Bronze!40}19.5&\cellcolor{Bronze!40}14.0&\cellcolor{Bronze!40}18.5&\cellcolor{Bronze!40}16.0&\cellcolor{Silver!60}47.8&\cellcolor{Silver!60}39.0&\cellcolor{Silver!60}13.0&\cellcolor{Silver!60}13.8&18.2& ~\mc{Gold}{0} & \mc{Silver}{4} & \mc{Bronze}{9} \\
Qwen3-8B &\cellcolor{Bronze!40}10.6&\cellcolor{Silver!60}12.7&11.5&\cellcolor{Bronze!40}7.1&\cellcolor{Bronze!40}11.9&\cellcolor{Silver!60}20.1&\cellcolor{Bronze!40}17.3&\cellcolor{Bronze!40}26.3&\cellcolor{Bronze!40}22.3&\cellcolor{Bronze!40}21.8&\cellcolor{Bronze!40}22.8&\cellcolor{Bronze!40}10.8&\cellcolor{Bronze!40}10.0&15.8& ~\mc{Gold}{0} & \mc{Silver}{2} & \mc{Bronze}{10} \\
Qwen2.5-VL-72B-Instruct &\cellcolor{Bronze!40}10.6&\cellcolor{Bronze!40}7.2&\cellcolor{Bronze!40}13.6&\cellcolor{Bronze!40}6.1&8.1&13.3&11.4&\cellcolor{Bronze!40}26.8&\cellcolor{Bronze!40}18.2&\cellcolor{Bronze!40}24.0&\cellcolor{Silver!60}28.5&\cellcolor{Silver!60}13.5&9.8&14.7& ~\mc{Gold}{0} & \mc{Silver}{2} & \mc{Bronze}{7} \\
LLaMA4-Scout-17B &\cellcolor{Bronze!40}9.7&\cellcolor{Bronze!40}9.5&\cellcolor{Bronze!40}13.1&5.4&\cellcolor{Bronze!40}10.4&\cellcolor{Silver!60}22.8&12.8&\cellcolor{Bronze!40}26.6&\cellcolor{Bronze!40}24.1&\cellcolor{Bronze!40}35.4&\cellcolor{Silver!60}34.5&8.1&6.4&16.8& ~\mc{Gold}{0} & \mc{Silver}{2} & \mc{Bronze}{7}\\
Qwen2.5-VL-32B-Instruct &\cellcolor{Bronze!40}9.9&\cellcolor{Bronze!40}8.2&\cellcolor{Bronze!40}16.5&\cellcolor{Bronze!40}6.9&\cellcolor{Bronze!40}10.0&\cellcolor{Bronze!40}15.3&\cellcolor{Bronze!40}14.4&\cellcolor{Bronze!40}22.5&\cellcolor{Bronze!40}22.4&\cellcolor{Bronze!40}28.1&\cellcolor{Silver!60}29.9&7.6&4.6&15.1& ~\mc{Gold}{0} & \mc{Silver}{1} & \mc{Bronze}{10} \\
InternVL3-38B-Instruct &\cellcolor{Bronze!40}8.9&\cellcolor{Bronze!40}7.8&12.3&\cellcolor{Bronze!40}6.1&8.3&14.0&10.6&\cellcolor{Bronze!40}24.1&\cellcolor{Bronze!40}20.4&\cellcolor{Bronze!40}27.5&\cellcolor{Bronze!40}24.8&8.2&6.8&13.8& ~\mc{Gold}{0} & \mc{Silver}{0} & \mc{Bronze}{7}\\
InternVL3-9B-Instruct &4.7&\cellcolor{Bronze!40}3.7&7.2&4.2&4.1&9.4&6.0&12.4&11.0&11.6&\cellcolor{Bronze!40}16.3&6.4&6.2&7.9& ~\mc{Gold}{0} & \mc{Silver}{0} & \mc{Bronze}{2}\\
Qwen2.5-VL-7B-Instruct &3.5&2.5&5.7&4.4&3.6&7.3&5.5&\cellcolor{Bronze!40}14.7&7.6&9.8&11.5&4.4&3.5&6.5& ~\mc{Gold}{0} & \mc{Silver}{0} & \mc{Bronze}{1}\\
Phi-4-multimodal &2.0&1.6&4.2&3.6&3.6&5.0&4.5&8.3&9.0&10.0&10.1&4.4&5.0&5.5& ~\mc{Gold}{0} & \mc{Silver}{0} & \mc{Bronze}{0}\\
DeepSeek-VL2 &1.8&0.5&2.5&3.4&3.4&5.0&3.2&5.6&4.8&7.3&6.4&5.0&3.9&4.0& ~\mc{Gold}{0} & \mc{Silver}{0} & \mc{Bronze}{0}\\
\bottomrule
\end{tabular}%
}
% \vspace{2mm}
% \begin{minipage}{\textwidth}
% \scriptsize
% * For \texttt{Kimi-K2-Thinking}, the inference timeout is set to 7200 s, and 9.58\% of cases exceed this limit.
% \end{minipage}
\caption*{\scriptsize *  For \texttt{Kimi-K2-Thinking}, the inference timeout is set to 2 hours, and 9.58\% of cases exceed this limit.}
\vspace{-8mm}
\end{table}

\section{Experiment}

% \subsection{Settings and Baselines}
%%%%%%%%%%%%%%%%%%%%%%%%%%%%%%%%%%%%%%
\subsection{Experimental Setup}

\textbf{Test Dataset.} 
To evaluate performance on challenging physics Olympiads, we constructed the HiPhO benchmark \citep{2025hipho}, the first High School Physics Olympiad benchmark covering the 13 most recent high school physics Olympiads from 2024 to 2025. These competitions range from international to regional levels and span 7 major types: IPhO, APhO, EuPhO, NBPhO, PanPhO, PanMechanics, and F=MA. The selection was based on both their global influence and the availability of human score distributions\footnote{CPhO, USAPhO, and APhO-2024 were excluded due to the lack of complete official contestant scores.}.

\textbf{Comparison Models.}
We compared against 33 representative models, including 11 closed-source and 22 open-source models, selected to reflect the current frontier of large language models:  
\vspace{-10pt}
\begin{itemize}[left=0pt]
    \item \textbf{Closed-source models:} GPT-5 \citep{GPT-5}, o3 \citep{o3_o4-mini}, o4-mini (high) \citep{o3_o4-mini}, o4-mini \citep{o3_o4-mini}, GPT-4o \citep{GPT-4o}, Gemini-2.5-Pro \citep{Gemini-2.5}, Gemini-2.5-Flash-Thinking \citep{Gemini-2.5}, Grok-4 \citep{Grok-4}, Claude-4-Sonnet-Thinking \citep{Claude-3.7-Sonnet}, Claude-4-Sonnet \citep{Claude-3.7-Sonnet}, Mistral-Medium-3 \citep{Mistral-Medium-3}.
    
    \item \textbf{Open-source models:} Intern-S1 \citep{Intern-S1}, InternVL3 Series \citep{InternVL3}, Qwen2.5-VL Series \citep{Qwen2.5-VL}, GLM-4.5V \citep{GLM-4.5V}, DeepSeek-VL2 \citep{DeepSeek-VL2}, LLaMA4-Scout-17B \citep{LLaMA4-Scout}, Phi-4-multimodal \citep{Phi-4-multimodal}, GPT-OSS-120B \citep{2025GPT-OSS}, Kimi-K2-Thinking \citep{2025Kimi-K2}, Kimi-K2-Instruct \citep{2025Kimi-K2}, DeepSeek-R1 \citep{guo2025deepseek-r1}, DeepSeek-V3 \citep{2024DeepSeek-V3}, Qwen3 Series \citep{yang2025qwen3}.  
\end{itemize}

\textbf{Model Configurations.}
All models were evaluated under a standardized inference setup following the HiPhO protocol. The temperature was fixed at $0.6$, and the maximum token limit was set as large as permitted by each model. For every problem, we conducted $8$ independent inference runs and computed the average score per problem. These averages were then aggregated across problems to yield the final exam score for each Olympiad.

\textbf{Evaluation Method.} We follow the evaluation protocol of the HiPhO benchmark \citep{2025hipho} and use \texttt{Gemini-2.5-Flash} as the grader. The benchmark integrates both \emph{answer-level} and \emph{step-level} grading based on official marking schemes, allowing partial credit for correct intermediate reasoning. For each problem, the model's final score is defined as the maximum of its answer-level and step-level scores, consistent with human grading: a correct final answer receives full credit, while an incorrect one can still earn partial points for valid intermediate steps. The overall exam score is obtained by summing the scores across all problems. Unlike conventional benchmarks that report accuracy, we use the \textit{exam score} as the evaluation metric, enabling direct comparison between model performance and official medal thresholds.

%%%%%%%%%%%%%%%%%%%%%%%%%%%%%%%%%%%%%%

\subsection{Evaluation on Physics Olympiads}

\paragraph{Excellent Single Model Performance.} Evaluation results are presented in Table~\ref{tab:hipho_all}. As noted above, the P1 model series is trained purely through reinforcement learning. We first demonstrate the excellent single-model performance to evaluate the effectiveness of our training strategy.
\begin{itemize}[left=0pt]
    \item \textbf{P1-235B-A22B} ranks alongside Gemini-2.5-Pro and Gemini-2.5-Flash-Thinking at the top of the medal table, earning 12 gold and 1 silver medal, and surpassing major closed-source models such as GPT-5 (11 gold), Grok-4 (10 gold), and Claude-4-Sonnet-Thinking (8 gold). Impressively, P1-235B-A22B scores 21.2 / 30 at the latest International Physics Olympiad (IPhO 2025), ranks Top 3 globally—behind only Gemini-2.5-Pro (22.7) and GPT-5 (22.3)—becoming the first and only open-source model to achieve gold-medal performance on IPhO 2025.

    \item \textbf{P1-30B-A3B} earned 8 gold, 4 silver, and 1 bronze medal, ranking third among existing open-source models—just behind the much larger Qwen3-235B-A22B-Thinking-2507 and DeepSeek-R1. It surpasses comparable models such as Qwen3-32B (7 gold, 6 silver) and Qwen3-30B-A3B-Thinking-2507 (6 gold, 6 silver, 1 bronze), demonstrating strong performance-to-scale efficiency.
\end{itemize}

\paragraph{Agentic Boost.} With the combination with the PhysicsMinions system, P1’s average performance improves from 35.9 to 38.4, reaching the overall Top-1 position across all models and outperforming leading closed-source models such as Gemini-2.5-Pro (37.7) and GPT-5 (37.4). P1-235B-A22B + PhysicsMinions also achieved new state-of-the-art results across four physics Olympiads. The joint configuration outperformed the best compared models on IPhO 2025 (23.2 vs. 22.7), slightly surpassed the record on APhO 2025 (28.0 vs. 27.9) and EuPhO 2024 (23.5 vs. 23.4), and secured a decisive advantage on PanMechanics 2025 (77.5 vs. 72.1). These results highlight how the integration of P1 with the multi-agent PhysicsMinions framework substantially enhances reasoning and problem-solving performance, showcasing the ``model + system'' paradigm for complex scientific reasoning.

%%%%%%%%%%%%%%%%%%%%%%%%%%%%%%%%%%%%%%

% \footnote{CPhO 2025 took place on Oct. 25th, 2025.}

\paragraph{Gold-Level Performance on CPhO 2025.}
We further evaluate \texttt{P1-235B-A22B} on the newly held 2025 Chinese Physics Olympiad (CPhO 2025), one of the most challenging physics Olympiads worldwide, renowned for its long multi-step reasoning problems. On the theoretical exam, \texttt{P1-235B-A22B} obtains a score of 227 / 320, assessed by human experts strictly following the official marking scheme. This score substantially exceeds the top-1 human medalist's 199, as shown in Table~\ref{tab:cpho2025}.
This demonstrates that our open-source model can attain gold-medal performance on CPhO 2025, marking an important milestone for open-source physics reasoning and showing that it can already match and even exceed elite human performance on some of the most challenging physics Olympiads.
% \vspace{-1mm}
\begin{table}[H]
    \centering
    % \vspace{-10pt}
    \caption{Performance comparison between P1-235B-A22B and top-1 human medalist on CPhO 2025.}
    \label{tab:cpho2025}
    \vspace{-1mm}
    \small
    \begin{tabular}{lcccccccc}
    \toprule
        Theoretical Problem & Q1 & Q2 & Q3 & Q4 & Q5 & Q6 & Q7 & Total Score  \\ \hline
        Full Mark & 45 & 45 & 45 & 45 & 45 & 50 & 45 & 320 \\
        Top-1 Human Medalist & 43 & 14 & 32 & 26 & 40 & 29 & 15 & 199 \\
        \rowcolor[HTML]{eee5f8} \textbf{P1-235B-A22B} & 35 & 25 & 39 & 21 & 31 & 39 & 37 & 227 \\
    \bottomrule
    \end{tabular}
    \vspace{-3mm}
\end{table}

% \clearpage
\section{Discussion}

\subsection{Generalizability of P1}
\begin{figure}[ht]
    \centering
    \subfloat{\includegraphics[width=0.48\textwidth]{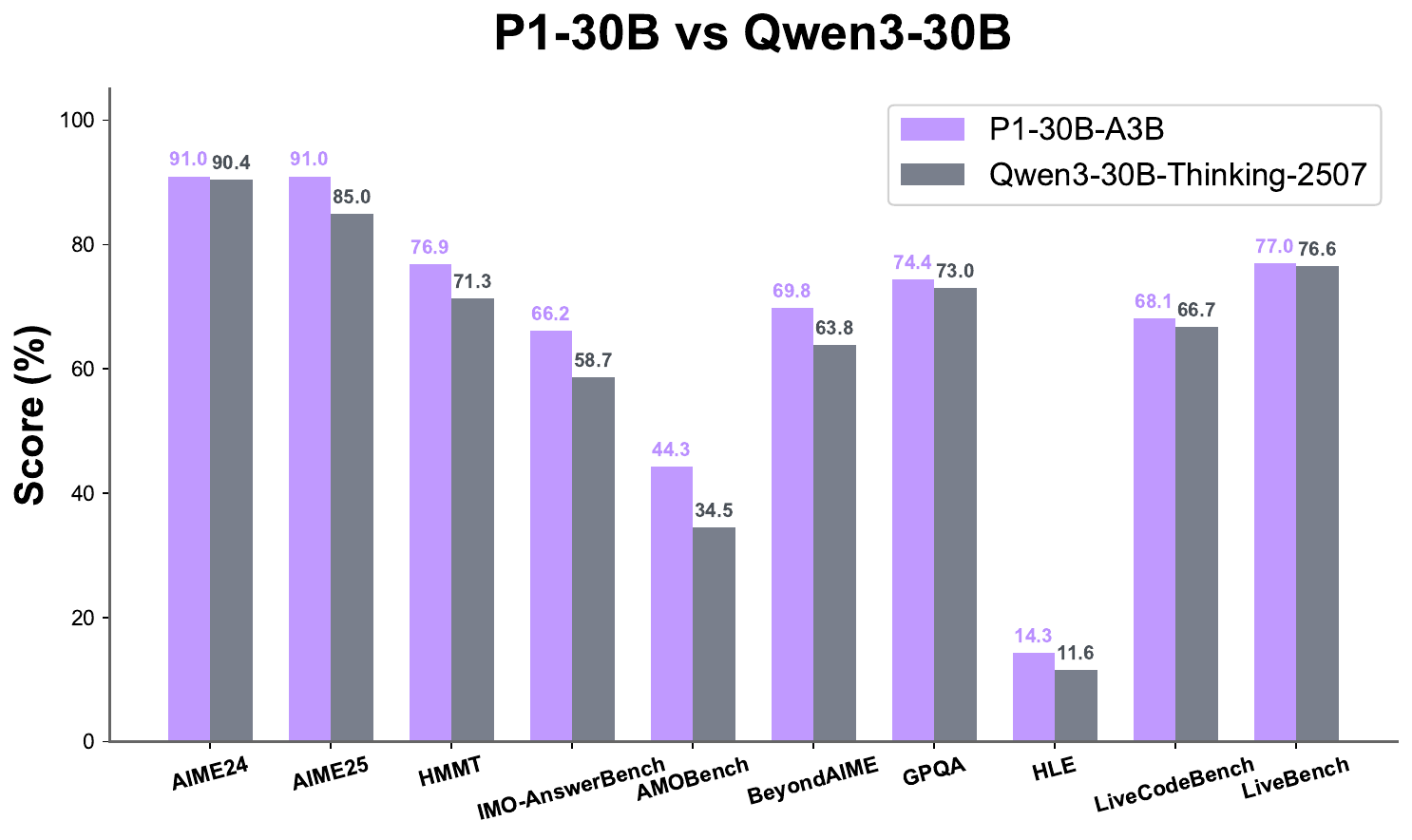}}
    \hfill
    \subfloat{\includegraphics[width=0.48\textwidth]{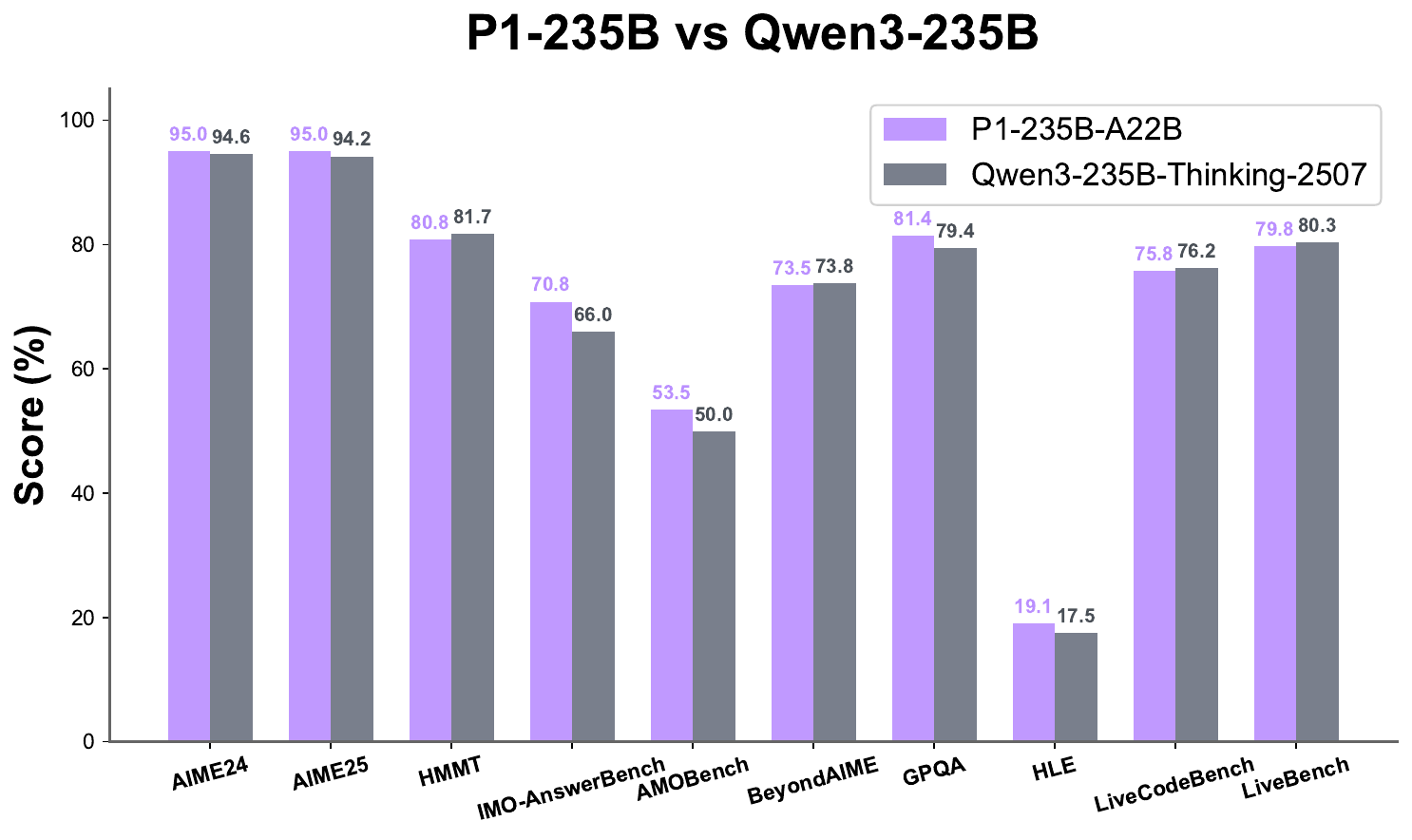}}
    
    \caption{Performance comparison between P1 models and base models on math and code datasets. P1 models present great general reasoning ability across mathematics, STEM, and coding domains.}
    \label{fig:general}
\end{figure}

We conduct post-training on a specialized dataset to enhance physics problem-solving abilities. Beyond domain-specific improvements, we further investigate: \emph{Does the P1 series model preserve, or even enhance, its general reasoning ability across mathematics, STEM, and coding domains?} To examine this, we compare the P1 series models with their respective base models across diverse benchmarks: six mathematics datasets (AIME24, AIME25, HMMT~\citep{matharena}, IMO-AnswerBench~\citep{imobench}, AMOBench~\citep{an2025amobench}, BeyondAIME~\citep{bytedance_seed_2025_beyondaime}), two STEM-oriented evaluations (GPQA~\citep{rein2024gpqa}, HLE~\citep{phan2025hle}), one coding benchmark (LiveCodeBench~\citep{jain2024livecodebench}), and a general reasoning task (LiveBench~\citep{white2024livebench}). Figure~\ref{fig:general} summarizes the results.

Notably, the P1 models demonstrate consistent advantages over their base counterparts: P1-30B-A3B outperforms Qwen3-30B-A3B-Thinking-2507 across all metrics, while P1-235B-A22B achieves superior performance to Qwen3-235B-A22B-Thinking-2507 in most categories (AIME24, AIME25, GPQA, HLE, IMO-AnswerBench, AMOBench, and overall average). This pattern highlights that P1 models not only maintain but enhance general reasoning abilities beyond the target domain—even on more challenging mathematics benchmarks (e.g., IMO-AnswerBench, AMOBench) that test advanced problem-solving skills.

These results suggest that domain-focused post-training can induce transferable improvements in general reasoning. We hypothesize two contributing factors. First, the optimization process refines reasoning trajectories in a way that transcends domain boundaries, enabling strategies beneficial for mathematics (including advanced subsets like IMO-style problems), STEM, and coding tasks to emerge. Second, the training dataset, while specialized, shares structural similarities with other domains—such as rigorous symbolic manipulation, multi-step logical deduction, and abstract problem modeling—that underpin general reasoning. Thus, the P1 models generalize effectively to neighboring domains and advanced sub-domains of mathematics, providing evidence that domain-focused post-training can simultaneously act as a general reasoning amplifier.

% \begin{table}[th]
% \centering
% \caption{Performance comparison between P1 models and base models on math and code datasets.}
% \label{tab:generalizability_comparison}
% \resizebox{\linewidth}{!}{
% \begin{tabular}{lccccccccccc}
% \toprule
% \multirow{2}{*}{\textbf{Model}} & \multicolumn{6}{c}{\textbf{Math}} &\multicolumn{2}{c}{\textbf{STEM}} & \multicolumn{1}{c}{\textbf{Coding}} & \multicolumn{1}{c}{\textbf{General}}  & \multirow{2}{*}{\textbf{Avg}} \\
% \cmidrule(lr){2-7} \cmidrule(lr){8-9} \cmidrule(lr){10-10} \cmidrule(lr){11-11}
% & \textbf{AIME24} & \textbf{AIME25} & \textbf{HMMT} & \textbf{IMO-AnswerBench} & \textbf{AMOBench} & \textbf{BeyondAIME} &  \textbf{GPQA} & \textbf{HLE} & \textbf{LiveCodeBench} & \textbf{LiveBench}  \\
% \midrule
% \textbf{Qwen3-30B-A3B-Thinking-2507}  & 90.4 & 85.0 & 71.3 & 58.7 & 34.5 & 63.8 & 73.0 & 11.6 & 66.7 & 76.6 & 63.2 \\
% \rowcolor[HTML]{eee5f8} \textbf{P1-30B-A3B}     & \textbf{91.0} & \textbf{91.0} & \textbf{76.9} & \textbf{66.2} & \textbf{44.3} & \textbf{69.8} & \textbf{74.4} & \textbf{14.3} & \textbf{68.1} & \textbf{77.0} & \textbf{67.3} \\
% \textbf{Qwen3-235B-A22B-Thinking-2507} & 94.6 & 94.2 & \textbf{81.7} & 66.0 & 50.0 & \textbf{73.8} & 79.4 & 17.5 & \textbf{76.2} & \textbf{80.3} & 71.4 \\
% \rowcolor[HTML]{eee5f8} \textbf{P1-235B-A22B}    & \textbf{95.0} & \textbf{95.0} & 80.8 & \textbf{70.8} & \textbf{53.5} & 73.5 & \textbf{81.4} & \textbf{19.1} & 75.8 & 79.8 & \textbf{72.5} \\
% \bottomrule
% \end{tabular}
% }
% \end{table}

\begin{figure}[ht]
    \centering
    \subfloat{\includegraphics[width=0.48\textwidth]{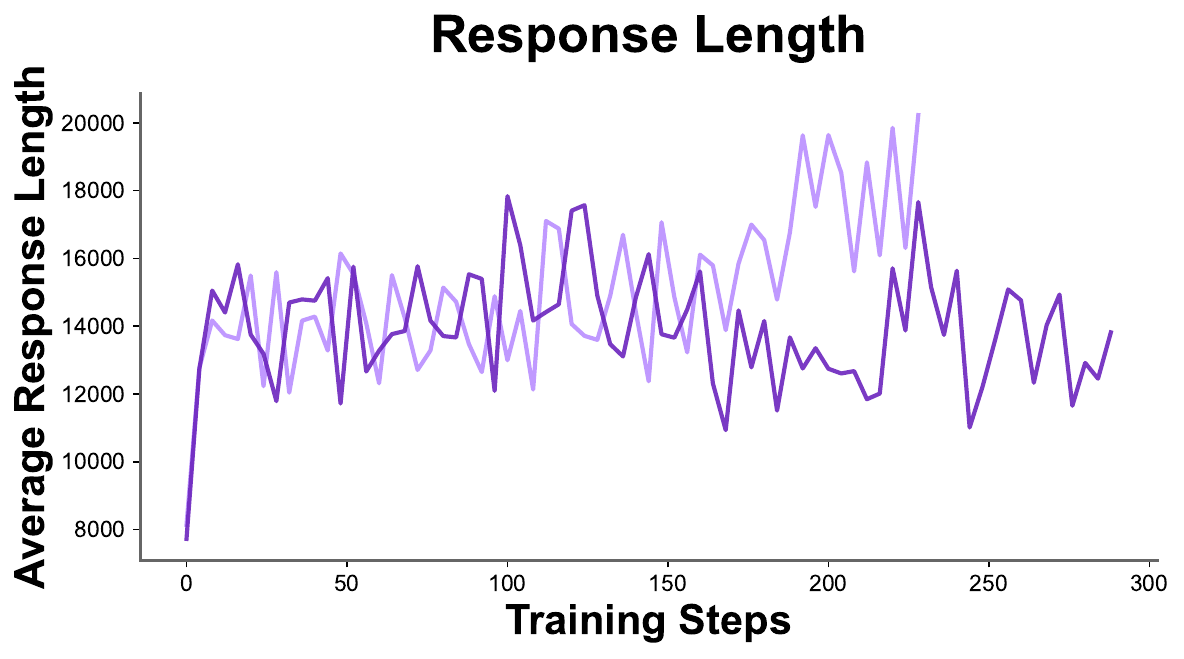}}
    \hfill
    \subfloat{\includegraphics[width=0.48\textwidth]{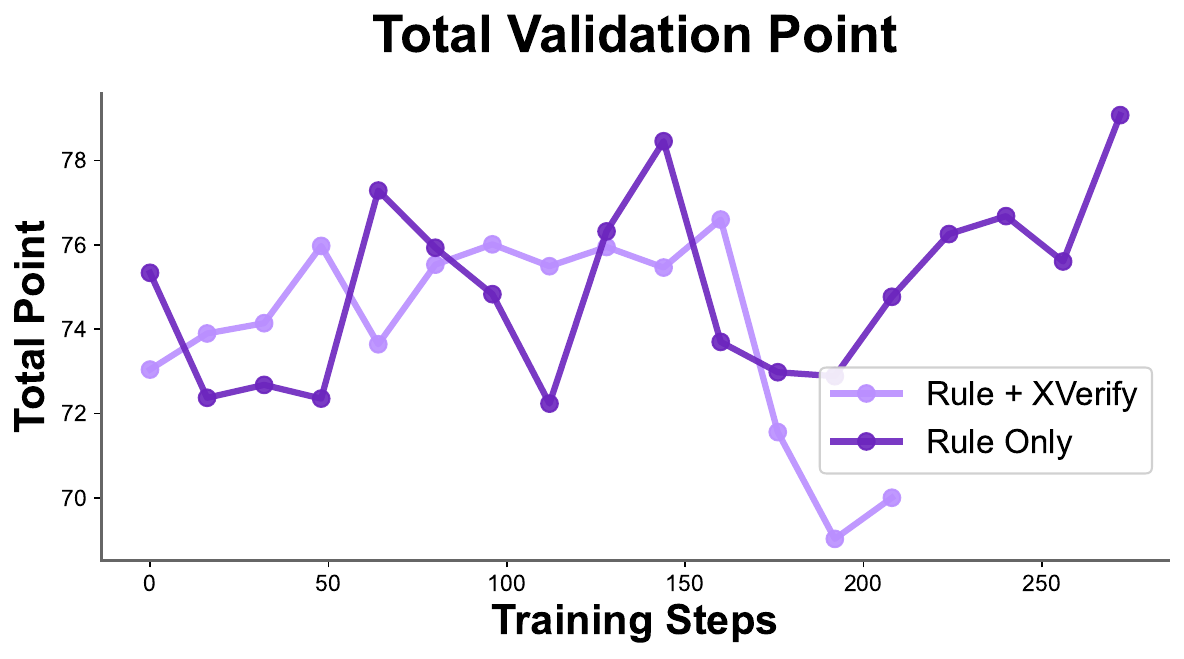}}
    
    \caption{Training dynamics differences when using an XVerify-like judge during training (P1-235B-A22B). The LHS shows the average response length on the training dataset. The RHS shows the total validation points across 5 representative competitions. We can observe that the inclusion of the XVerify judge even exerts negative effects on the post-training process, as indicated by a consistent increase in response length alongside degraded validation performance.}
    \label{fig:xverify}
\end{figure}

\subsection{Rule-based vs Model-based Verifier for Training}\label{sec:xverify}

As stated in Section~\ref{train_dyn}, due to the difficulty of implementing a comprehensive and fully correct verification mechanism purely based on rules, we design a hybrid verifier that integrates both rule-based and model-based reasoning to achieve broader coverage. However, we found that applying a model-based verifier directly in the post-training process can be risky. Figure~\ref{fig:xverify} compares the training dynamics with and without a model-based verifier. We observe that the variant employing the model-based verifier exhibits an explosive increase in response length, yet its validation performance deteriorates compared to the setup using only a rule-based verifier.

We attribute this phenomenon to two primary causes.  
(1) The model-based verifier is susceptible to being \emph{hacked} by the policy model. Since the verifier itself is a language model, it may develop unintended biases, favoring atypical response patterns such as overly verbose or stylistically peculiar answers, which can be exploited by the policy model to obtain artificially high rewards.  
(2) The negative impact of \emph{false positives} (i.e., incorrectly rewarding wrong answers) is substantially more detrimental than that of \emph{false negatives} (i.e., missing some correct responses). In our observations, while the rule-based verifier fails to recognize certain valid answers, the model-based verifier expands the recognition scope to include more potentially correct samples—but at the cost of introducing incorrect judgments.

This trade-off can be understood in terms of \textbf{verification precision} and \textbf{verification recall}. The rule-based verifier typically offers high precision but limited recall, while the model-based verifier increases recall at the expense of precision. During reinforcement learning, this imbalance can easily destabilize optimization: a few high-reward false positives can dominate the learning signal, leading the policy toward degenerate solution patterns.

We emphasize that raising this issue is not to deny the value of model-based verifiers. Both verification precision and recall are critical for successful post-training. The key lies in ensuring sufficiently high precision before attempting to expand recall. Only when the model-based verifier achieves a stable and reliable judgment boundary can its inclusion provide net positive benefits. Therefore, this phenomenon highlights an urgent need for developing \emph{more robust and calibrated model-based verifiers}, capable of maintaining both correctness and coverage under the dynamics of RL-based post-training.

\section{Conclusion}
We release the first family of open-source models capable of mastering Olympiad-level physics problems, P1. P1 models achieve great performance on physics reasoning tasks, even reaching goal medal performance on the latest Olympiad competitions. These achievements are made possible by the combination of train-time scaling via RL post-training and test-time scaling via agentic framework. Besides physics, P1 models also present great performance on other reasoning tasks like math and coding, showing the great generalibility of P1 series. Looking forward, P1's success in mastering Olympiad-level physics represents a key milestone toward LLMs that can assist or pioneer real-world physics research: if models can rigorously solve well-defined problems grounded in natural laws, they may eventually contribute to exploring uncharted scientific frontiers.

\section{Acknowlegement}
This work is supported by Shanghai AI Laboratory. 
We would like to extend our special thanks to the developers and maintainers of the following open-source projects, which have been critical to the implementation of this work. This includes Qwen3~\citep{yang2025qwen3}, which provided the foundational base models for our research; slime~\citep{slime_github}, whose innovative framework enabled efficient reinforcement learning in our training pipeline; and verl~\citep{sheng2024verl}, which offered a versatile reinforcement learning framework to support model training. We also thank sglang~\citep{zheng2024sglang} for its efficient infrastructure for LLM serving and inference, and Megatron-LM~\citep{shoeybi2019megatron} for providing the large-scale model training framework.

% \newpage
\bibliography{main}

@article{qiu2025physicsagent,
  title={Physics supernova: Ai agent matches elite gold medalists at ipho 2025},
  author={Qiu, Jiahao and Shi, Jingzhe and Juan, Xinzhe and Zhao, Zelin and Geng, Jiayi and Liu, Shilong and Wang, Hongru and Wu, Sanfeng and Wang, Mengdi},
  journal={arXiv preprint arXiv:2509.01659},
  year={2025}
}

@article{zhang2024llmsci_survey,
  title={A comprehensive survey of scientific large language models and their applications in scientific discovery},
  author={Zhang, Yu and Chen, Xiusi and Jin, Bowen and Wang, Sheng and Ji, Shuiwang and Wang, Wei and Han, Jiawei},
  journal={arXiv preprint arXiv:2406.10833},
  year={2024}
}

@article{guo2025deepseek-r1,
  title={Deepseek-r1: Incentivizing reasoning capability in llms via reinforcement learning},
  author={Guo, Daya and Yang, Dejian and Zhang, Haowei and Song, Junxiao and Zhang, Ruoyu and Xu, Runxin and Zhu, Qihao and Ma, Shirong and Wang, Peiyi and Bi, Xiao and others},
  journal={arXiv preprint arXiv:2501.12948},
  year={2025}
}

@article{yang2025qwen3,
  title={Qwen3 technical report},
  author={Yang, An and Li, Anfeng and Yang, Baosong and Zhang, Beichen and Hui, Binyuan and Zheng, Bo and Yu, Bowen and Gao, Chang and Huang, Chengen and Lv, Chenxu and others},
  journal={arXiv preprint arXiv:2505.09388},
  year={2025}
}

@article{gibney2025deepmind_sci,
  title={DeepMind unveils ‘spectacular’general-purpose science AI},
  author={Gibney, Elizabeth},
  journal={Nature},
  volume={641},
  number={8064},
  pages={827--828},
  year={2025},
  publisher={Nature}
}

@article{oh2025deepmind_discover_rl,
  title={Discovering state-of-the-art reinforcement learning algorithms},
  author={Oh, Junhyuk and Farquhar, Greg and Kemaev, Iurii and Calian, Dan A and Hessel, Matteo and Zintgraf, Luisa and Singh, Satinder and van Hasselt, Hado and Silver, David},
  journal={Nature},
  pages={1--2},
  year={2025},
  publisher={Nature Publishing Group UK London}
}

@article{zheng2025llm-discover,
  title={Large language models for scientific discovery in molecular property prediction},
  author={Zheng, Yizhen and Koh, Huan Yee and Ju, Jiaxin and Nguyen, Anh TN and May, Lauren T and Webb, Geoffrey I and Pan, Shirui},
  journal={Nature Machine Intelligence},
  pages={1--11},
  year={2025},
  publisher={Nature Publishing Group UK London}
}

@book{sutton1998rl-basis,
  title={Reinforcement learning: An introduction},
  author={Sutton, Richard S and Barto, Andrew G and others},
  volume={1},
  number={1},
  year={1998},
  publisher={MIT press Cambridge}
}

@article{cui2025prime-rl,
  title={Process reinforcement through implicit rewards},
  author={Cui, Ganqu and Yuan, Lifan and Wang, Zefan and Wang, Hanbin and Li, Wendi and He, Bingxiang and Fan, Yuchen and Yu, Tianyu and Xu, Qixin and Chen, Weize and others},
  journal={arXiv preprint arXiv:2502.01456},
  year={2025}
}

@article{zheng2025scaling,
  title={Scaling physical reasoning with the physics dataset},
  author={Zheng, Shenghe and Cheng, Qianjia and Yao, Junchi and Wu, Mengsong and He, Haonan and Ding, Ning and Cheng, Yu and Hu, Shuyue and Bai, Lei and Zhou, Dongzhan and others},
  journal={arXiv preprint arXiv:2506.00022},
  year={2025}
}

@article{wen2025reinforcement,
  title={Reinforcement Learning with Verifiable Rewards Implicitly Incentivizes Correct Reasoning in Base LLMs},
  author={Wen, Xumeng and Liu, Zihan and Zheng, Shun and Xu, Zhijian and Ye, Shengyu and Wu, Zhirong and Liang, Xiao and Wang, Yang and Li, Junjie and Miao, Ziming and others},
  journal={arXiv preprint arXiv:2506.14245},
  year={2025}
}

@article{sutton1999pg-sutton,
  title={Policy gradient methods for reinforcement learning with function approximation},
  author={Sutton, Richard S and McAllester, David and Singh, Satinder and Mansour, Yishay},
  journal={Advances in neural information processing systems},
  volume={12},
  year={1999}
}

@article{zheng2025gspo,
  title={Group sequence policy optimization},
  author={Zheng, Chujie and Liu, Shixuan and Li, Mingze and Chen, Xiong-Hui and Yu, Bowen and Gao, Chang and Dang, Kai and Liu, Yuqiong and Men, Rui and Yang, An and others},
  journal={arXiv preprint arXiv:2507.18071},
  year={2025}
}

@article{shao2024grpo,
  title={Deepseekmath: Pushing the limits of mathematical reasoning in open language models},
  author={Shao, Zhihong and Wang, Peiyi and Zhu, Qihao and Xu, Runxin and Song, Junxiao and Bi, Xiao and Zhang, Haowei and Zhang, Mingchuan and Li, YK and Wu, Yang and others},
  journal={arXiv preprint arXiv:2402.03300},
  year={2024}
}

@article{yu2025dapo,
  title={Dapo: An open-source llm reinforcement learning system at scale},
  author={Yu, Qiying and Zhang, Zheng and Zhu, Ruofei and Yuan, Yufeng and Zuo, Xiaochen and Yue, Yu and Dai, Weinan and Fan, Tiantian and Liu, Gaohong and Liu, Lingjun and others},
  journal={arXiv preprint arXiv:2503.14476},
  year={2025}
}

@article{liu2025drgrpo,
  title={Understanding r1-zero-like training: A critical perspective},
  author={Liu, Zichen and Chen, Changyu and Li, Wenjun and Qi, Penghui and Pang, Tianyu and Du, Chao and Lee, Wee Sun and Lin, Min},
  journal={arXiv preprint arXiv:2503.20783},
  year={2025}
}

@article{chen2025xverify,
  title={xverify: Efficient answer verifier for reasoning model evaluations},
  author={Chen, Ding and Yu, Qingchen and Wang, Pengyuan and Zhang, Wentao and Tang, Bo and Xiong, Feiyu and Li, Xinchi and Yang, Minchuan and Li, Zhiyu},
  journal={arXiv preprint arXiv:2504.10481},
  year={2025}
}

@article{cui2025entropy,
  title={The entropy mechanism of reinforcement learning for reasoning language models},
  author={Cui, Ganqu and Zhang, Yuchen and Chen, Jiacheng and Yuan, Lifan and Wang, Zhi and Zuo, Yuxin and Li, Haozhan and Fan, Yuchen and Chen, Huayu and Chen, Weize and others},
  journal={arXiv preprint arXiv:2505.22617},
  year={2025}
}

@article{meurer2017sympy,
  title={SymPy: symbolic computing in Python},
  author={Meurer, Aaron and Smith, Christopher P and Paprocki, Mateusz and {\v{C}}ert{\'\i}k, Ond{\v{r}}ej and Kirpichev, Sergey B and Rocklin, Matthew and Kumar, AMiT and Ivanov, Sergiu and Moore, Jason K and Singh, Sartaj and others},
  journal={PeerJ Computer Science},
  volume={3},
  pages={e103},
  year={2017},
  publisher={PeerJ Inc.}
}

@software{math-verify,
author = {Kydlíček, Hynek},
license = {Apache-2.0},
title = {{Math-Verify: Math Verification Library}},
url = {https://github.com/huggingface/math-verify},
version = {0.6.1}
}

@article{zhang2025survey-reasoning,
  title={A survey of reinforcement learning for large reasoning models},
  author={Zhang, Kaiyan and Zuo, Yuxin and He, Bingxiang and Sun, Youbang and Liu, Runze and Jiang, Che and Fan, Yuchen and Tian, Kai and Jia, Guoli and Li, Pengfei and others},
  journal={arXiv preprint arXiv:2509.08827},
  year={2025}
}

@misc{deepscaler2025,
  title={DeepScaleR: Surpassing O1-Preview with a 1.5B Model by Scaling RL},
  author={Michael Luo and Sijun Tan and Justin Wong and Xiaoxiang Shi and William Y. Tang and Manan Roongta and Colin Cai and Jeffrey Luo and Li Erran Li and Raluca Ada Popa and Ion Stoica},
  year={2025},
  howpublished={\url{https://pretty-radio-b75.notion.site/DeepScaleR-Surpassing-O1-Preview-with-a-1-5B-Model-by-Scaling-RL-19681902c1468005bed8ca303013a4e2}},
  note={Notion Blog},
  year={2025}
}

@misc{yao2025tis,
  title = {Your Efficient RL Framework Secretly Brings You Off-Policy RL Training},
  url = {https://fengyao.notion.site/off-policy-rl},
  author = {Yao, Feng and Liu, Liyuan and Zhang, Dinghuai and Dong, Chengyu and Shang, Jingbo and Gao, Jianfeng},
  journal = {Feng Yao's Notion},
  year = {2025},
  month = aug,
}

@article{jiacai2025speed,
  title={When Speed Kills Stability: Demystifying RL Collapse from the Training-Inference Mismatch},
  author={Liu, Jiacai and Li, Yingru and Fu, Yuqian and Wang, Jiawei and Liu, Qian and Shen, Yu},
  journal={Notion Blog},
  year={2025}
}

@inproceedings{imobench,
    title = "Towards Robust Mathematical Reasoning",
    author  = {Thang Luong and Dawsen Hwang and Hoang H. Nguyen and Golnaz Ghiasi and Yuri Chervonyi and Insuk Seo and Junsu Kim and Garrett Bingham and Jonathan Lee and Swaroop Mishra and Alex Zhai and Clara Huiyi Hu and Henryk Michalewski and Jimin Kim and Jeonghyun Ahn and Junhwi Bae and Xingyou Song and Trieu H. Trinh and Quoc V. Le and Junehyuk Jung},
    booktitle = "Proceedings of the 2025 Conference on Empirical Methods in Natural Language Processing",
    year = "2025",
    url = "https://aclanthology.org/2025.emnlp-main.1794/",
}

@misc{bytedance_seed_2025_beyondaime,
  author       = {ByteDance-Seed},
  title        = {BeyondAIME: Advancing Math Reasoning Evaluation Beyond High School Olympiads},
  year         = {2025},
  publisher    = {Hugging Face},
  journal      = {Hugging Face repository},
  howpublished = {\url{[https://huggingface.co/datasets/ByteDance-Seed/BeyondAIME](https://huggingface.co/datasets/ByteDance-Seed/BeyondAIME)}},
}

@misc{an2025amobench,
      title={AMO-Bench: Large Language Models Still Struggle in High School Math Competitions}, 
      author={Shengnan An and Xunliang Cai and Xuezhi Cao and Xiaoyu Li and Yehao Lin and Junlin Liu and Xinxuan Lv and Dan Ma and Xuanlin Wang and Ziwen Wang and Shuang Zhou},
      year={2025},
      eprint={2510.26768},
      archivePrefix={arXiv},
      primaryClass={cs.CL},
      url={https://arxiv.org/abs/2510.26768}, 
}

@misc{matharena,
  title = {MathArena: Evaluating LLMs on Uncontaminated Math Competitions},
  author = {Mislav Balunović and Jasper Dekoninck and Ivo Petrov and Nikola Jovanović and Martin Vechev},
  copyright = {MIT},
  url = {https://matharena.ai/},
  publisher = {SRI Lab, ETH Zurich},
  month = feb,
  year = {2025},
}

@inproceedings{rein2024gpqa,
  title={Gpqa: A graduate-level google-proof q\&a benchmark},
  author={Rein, David and Hou, Betty Li and Stickland, Asa Cooper and Petty, Jackson and Pang, Richard Yuanzhe and Dirani, Julien and Michael, Julian and Bowman, Samuel R},
  booktitle={First Conference on Language Modeling},
  year={2024}
}

@article{phan2025hle,
  title={Humanity's last exam},
  author={Phan, Long and Gatti, Alice and Han, Ziwen and Li, Nathaniel and Hu, Josephina and Zhang, Hugh and Zhang, Chen Bo Calvin and Shaaban, Mohamed and Ling, John and Shi, Sean and others},
  journal={arXiv preprint arXiv:2501.14249},
  year={2025}
}

@article{jain2024livecodebench,
  title={Livecodebench: Holistic and contamination free evaluation of large language models for code},
  author={Jain, Naman and Han, King and Gu, Alex and Li, Wen-Ding and Yan, Fanjia and Zhang, Tianjun and Wang, Sida and Solar-Lezama, Armando and Sen, Koushik and Stoica, Ion},
  journal={arXiv preprint arXiv:2403.07974},
  year={2024}
}

@article{white2024livebench,
  title={Livebench: A challenging, contamination-free llm benchmark},
  author={White, Colin and Dooley, Samuel and Roberts, Manley and Pal, Arka and Feuer, Ben and Jain, Siddhartha and Shwartz-Ziv, Ravid and Jain, Neel and Saifullah, Khalid and Naidu, Siddartha and others},
  journal={arXiv preprint arXiv:2406.19314},
  volume={4},
  year={2024}
}

@misc{slime_github,
  author       = {Zilin Zhu and Chengxing Xie and Xin Lv and slime Contributors},
  title        = {slime: An LLM post-training framework for RL Scaling},
  year         = {2025},
  howpublished = {\url{https://github.com/THUDM/slime}},
  note         = {GitHub repository. Corresponding author: Xin Lv},
  urldate      = {2025-06-19}
}

@article{merry2021fsdp,
  title={Fully Sharded Data Parallel: Reducing Memory Usage for Large Model Training},
  author={Merry, Andrew and Rajbhandari, Samyam and Shoeybi, Mohammad and Puri, Ramesh and Fung, Paul and Anandkumar, Anima and Catanzaro, Bryan},
  journal={PyTorch Developer Blog},
  year={2021},
  url={https://pytorch.org/blog/introducing-pytorch-fully-sharded-data-parallel-api/}
}

@inproceedings{kwon2023vllm,
  title={Efficient Memory Management for Large Language Model Serving with PagedAttention},
  author={Woosuk Kwon and Zhuohan Li and Siyuan Zhuang and Ying Sheng and Lianmin Zheng and Cody Hao Yu and Joseph E. Gonzalez and Hao Zhang and Ion Stoica},
  booktitle={Proceedings of the ACM SIGOPS 29th Symposium on Operating Systems Principles},
  year={2023}
}

@article{sheng2024verl,
  title   = {HybridFlow: A Flexible and Efficient RLHF Framework},
  author  = {Guangming Sheng and Chi Zhang and Zilingfeng Ye and Xibin Wu and Wang Zhang and Ru Zhang and Yanghua Peng and Haibin Lin and Chuan Wu},
  year    = {2024},
  journal = {arXiv preprint arXiv: 2409.19256}
}

@article{zheng2024sglang,
  title={Sglang: Efficient execution of structured language model programs},
  author={Zheng, Lianmin and Yin, Liangsheng and Xie, Zhiqiang and Sun, Chuyue Livia and Huang, Jeff and Yu, Cody Hao and Cao, Shiyi and Kozyrakis, Christos and Stoica, Ion and Gonzalez, Joseph E and others},
  journal={Advances in neural information processing systems},
  volume={37},
  pages={62557--62583},
  year={2024}
}

@article{shoeybi2019megatron,
  title={Megatron-lm: Training multi-billion parameter language models using model parallelism},
  author={Shoeybi, Mohammad and Patwary, Mostofa and Puri, Raul and LeGresley, Patrick and Casper, Jared and Catanzaro, Bryan},
  journal={arXiv preprint arXiv:1909.08053},
  year={2019}
}

@article{2025hipho,
  title={HiPhO: How Far Are (M)LLMs from Humans in the Latest High School Physics Olympiad Benchmark?},
  author={Yu, Fangchen and Wan, Haiyuan and Cheng, Qianjia and Zhang, Yuchen and Chen, Jiacheng and Han, Fujun and Wu, Yulun and Yao, Junchi and Hu, Ruilizhen and Ding, Ning and Cheng, Yu and Chen, Tao and Bai, Lei and Zhou, Dongzhan and Luo, Yun and Cui, Ganqu and Ye, Peng},
  journal={arXiv preprint arXiv:2509.07894},
  year={2025}
}

@misc{physicsminions,
      title={PhysicsMinions: Winning Gold Medals in the Latest Physics Olympiads with a Coevolutionary Multimodal Multi-Agent System}, 
      author={Fangchen Yu and Junchi Yao and Ziyi Wang and Haiyuan Wan and Youling Huang and Bo Zhang and Shuyue Hu and Dongzhan Zhou and Ning Ding and Ganqu Cui and Lei Bai and Wanli Ouyang and Peng Ye},
      year={2025},
      eprint={2509.24855},
      archivePrefix={arXiv},
      primaryClass={cs.AI},
      url={https://arxiv.org/abs/2509.24855}, 
}

@article{Gemini-2.5,
  title={Gemini 2.5: Pushing the frontier with advanced reasoning, multimodality, long context, and next generation agentic capabilities},
  author={Comanici, Gheorghe and Bieber, Eric and Schaekermann, Mike and Pasupat, Ice and Sachdeva, Noveen and Dhillon, Inderjit and Blistein, Marcel and Ram, Ori and Zhang, Dan and Rosen, Evan and others},
  journal={arXiv preprint arXiv:2507.06261},
  year={2025}
}

@misc{GPT-5,
  title = {GPT-5 System Card},
  url = {https://openai.com/index/gpt-5-system-card/},
  author = {OpenAI},
}

@misc{o3_o4-mini,
  title = {OpenAI o3 and o4-mini System Card},
  url = {https://openai.com/index/introducing-o3-and-o4-mini/},
  author = {OpenAI},
}

@misc{GPT-4o,
  title = {GPT-4o System Card},
  url = {https://openai.com/index/gpt-4o-system-card/},
  author = {OpenAI},
}

@misc{Grok-4,
  title = {Grok 4 System Card},
  url = {https://x.ai/grok},
  author = {xAI},
}

@misc{Claude-3.7-Sonnet,
  title = {Claude 3.7 Sonnet System Card},
  url = {https://www.anthropic.com/news/visible-extended-thinking},
  author = {Anthropic},
}

@misc{Mistral-Medium-3,
  title = {Mistral-Medium-3 System Card},
  url = {https://mistral.ai/news/mistral-medium-3},
  author = {Mistral},
}

@misc{Intern-S1,
      title={Intern-S1: A Scientific Multimodal Foundation Model},
      author={Lei Bai and Zhongrui Cai and Maosong Cao and Weihan Cao and Chiyu Chen and Haojiong Chen and Kai Chen and Pengcheng Chen and Ying Chen and Yongkang Chen and Yu Cheng and Yu Cheng and Pei Chu and Tao Chu and Erfei Cui and Ganqu Cui and Long Cui and Ziyun Cui and Nianchen Deng and Ning Ding and Nanqin Dong and Peijie Dong and Shihan Dou and Sinan Du and Haodong Duan and Caihua Fan and Ben Gao and Changjiang Gao and Jianfei Gao and Songyang Gao and Yang Gao and Zhangwei Gao and Jiaye Ge and Qiming Ge and Lixin Gu and Yuzhe Gu and Aijia Guo and Qipeng Guo and Xu Guo and Conghui He and Junjun He and Yili Hong and Siyuan Hou and Caiyu Hu and Hanglei Hu and Jucheng Hu and Ming Hu and Zhouqi Hua and Haian Huang and Junhao Huang and Xu Huang and Zixian Huang and Zhe Jiang and Lingkai Kong and Linyang Li and Peiji Li and Pengze Li and Shuaibin Li and Tianbin Li and Wei Li and Yuqiang Li and Dahua Lin and Junyao Lin and Tianyi Lin and Zhishan Lin and Hongwei Liu and Jiangning Liu and Jiyao Liu and Junnan Liu and Kai Liu and Kaiwen Liu and Kuikun Liu and Shichun Liu and Shudong Liu and Wei Liu and Xinyao Liu and Yuhong Liu and Zhan Liu and Yinquan Lu and Haijun Lv and Hongxia Lv and Huijie Lv and Qidang Lv and Ying Lv and Chengqi Lyu and Chenglong Ma and Jianpeng Ma and Ren Ma and Runmin Ma and Runyuan Ma and Xinzhu Ma and Yichuan Ma and Zihan Ma and Sixuan Mi and Junzhi Ning and Wenchang Ning and Xinle Pang and Jiahui Peng and Runyu Peng and Yu Qiao and Jiantao Qiu and Xiaoye Qu and Yuan Qu and Yuchen Ren and Fukai Shang and Wenqi Shao and Junhao Shen and Shuaike Shen and Chunfeng Song and Demin Song and Diping Song and Chenlin Su and Weijie Su and Weigao Sun and Yu Sun and Qian Tan and Cheng Tang and Huanze Tang and Kexian Tang and Shixiang Tang and Jian Tong and Aoran Wang and Bin Wang and Dong Wang and Lintao Wang and Rui Wang and Weiyun Wang and Wenhai Wang and Yi Wang and Ziyi Wang and Ling-I Wu and Wen Wu and Yue Wu and Zijian Wu and Linchen Xiao and Shuhao Xing and Chao Xu and Huihui Xu and Jun Xu and Ruiliang Xu and Wanghan Xu and GanLin Yang and Yuming Yang and Haochen Ye and Jin Ye and Shenglong Ye and Jia Yu and Jiashuo Yu and Jing Yu and Fei Yuan and Bo Zhang and Chao Zhang and Chen Zhang and Hongjie Zhang and Jin Zhang and Qiaosheng Zhang and Qiuyinzhe Zhang and Songyang Zhang and Taolin Zhang and Wenlong Zhang and Wenwei Zhang and Yechen Zhang and Ziyang Zhang and Haiteng Zhao and Qian Zhao and Xiangyu Zhao and Xiangyu Zhao and Bowen Zhou and Dongzhan Zhou and Peiheng Zhou and Yuhao Zhou and Yunhua Zhou and Dongsheng Zhu and Lin Zhu and Yicheng Zou},
      year={2025},
      eprint={2508.15763},
      archivePrefix={arXiv},
      primaryClass={cs.LG},
      url={https://arxiv.org/abs/2508.15763},
}

@article{InternVL3,
  title={Internvl3: Exploring advanced training and test-time recipes for open-source multimodal models},
  author={Zhu, Jinguo and Wang, Weiyun and Chen, Zhe and Liu, Zhaoyang and Ye, Shenglong and Gu, Lixin and Tian, Hao and Duan, Yuchen and Su, Weijie and Shao, Jie and others},
  journal={arXiv preprint arXiv:2504.10479},
  year={2025}
}

@article{Qwen2.5-VL,
  title={Qwen2.5-vl technical report},
  author={Bai, Shuai and Chen, Keqin and Liu, Xuejing and Wang, Jialin and Ge, Wenbin and Song, Sibo and Dang, Kai and Wang, Peng and Wang, Shijie and Tang, Jun and others},
  journal={arXiv preprint arXiv:2502.13923},
  year={2025}
}

@misc{GLM-4.5V,
      title={GLM-4.5V and GLM-4.1V-Thinking: Towards Versatile Multimodal Reasoning with Scalable Reinforcement Learning}, 
      author={GLM-V Team and Wenyi Hong and Wenmeng Yu and Xiaotao Gu and Guo Wang and Guobing Gan and Haomiao Tang and Jiale Cheng and Ji Qi and Junhui Ji and Lihang Pan and Shuaiqi Duan and Weihan Wang and Yan Wang and Yean Cheng and Zehai He and Zhe Su and Zhen Yang and Ziyang Pan and Aohan Zeng and Baoxu Wang and Bin Chen and Boyan Shi and Changyu Pang and Chenhui Zhang and Da Yin and Fan Yang and Guoqing Chen and Jiazheng Xu and Jiale Zhu and Jiali Chen and Jing Chen and Jinhao Chen and Jinghao Lin and Jinjiang Wang and Junjie Chen and Leqi Lei and Letian Gong and Leyi Pan and Mingdao Liu and Mingde Xu and Mingzhi Zhang and Qinkai Zheng and Sheng Yang and Shi Zhong and Shiyu Huang and Shuyuan Zhao and Siyan Xue and Shangqin Tu and Shengbiao Meng and Tianshu Zhang and Tianwei Luo and Tianxiang Hao and Tianyu Tong and Wenkai Li and Wei Jia and Xiao Liu and Xiaohan Zhang and Xin Lyu and Xinyue Fan and Xuancheng Huang and Yanling Wang and Yadong Xue and Yanfeng Wang and Yanzi Wang and Yifan An and Yifan Du and Yiming Shi and Yiheng Huang and Yilin Niu and Yuan Wang and Yuanchang Yue and Yuchen Li and Yutao Zhang and Yuting Wang and Yu Wang and Yuxuan Zhang and Zhao Xue and Zhenyu Hou and Zhengxiao Du and Zihan Wang and Peng Zhang and Debing Liu and Bin Xu and Juanzi Li and Minlie Huang and Yuxiao Dong and Jie Tang},
      year={2025},
      eprint={2507.01006},
      archivePrefix={arXiv},
      primaryClass={cs.CV},
      url={https://arxiv.org/abs/2507.01006}, 
}

@article{DeepSeek-VL2,
  title={Deepseek-vl2: Mixture-of-experts vision-language models for advanced multimodal understanding},
  author={Wu, Zhiyu and Chen, Xiaokang and Pan, Zizheng and Liu, Xingchao and Liu, Wen and Dai, Damai and Gao, Huazuo and Ma, Yiyang and Wu, Chengyue and Wang, Bingxuan and others},
  journal={arXiv preprint arXiv:2412.10302},
  year={2024}
}

@misc{LLaMA4-Scout,
  title = {Llama 4 System Card},
  url = {https://www.llama.com/docs/model-cards-and-prompt-formats/llama4/},
  author = {Meta},
}

@article{Phi-4-multimodal,
  title={Phi-4-mini technical report: Compact yet powerful multimodal language models via mixture-of-loras},
  author={Abouelenin, Abdelrahman and Ashfaq, Atabak and Atkinson, Adam and Awadalla, Hany and Bach, Nguyen and Bao, Jianmin and Benhaim, Alon and Cai, Martin and Chaudhary, Vishrav and Chen, Congcong and others},
  journal={arXiv preprint arXiv:2503.01743},
  year={2025}
}

@article{2025GPT-OSS,
  title={gpt-oss-120b \& gpt-oss-20b Model Card},
  author={Agarwal, Sandhini and Ahmad, Lama and Ai, Jason and Altman, Sam and Applebaum, Andy and Arbus, Edwin and Arora, Rahul K and Bai, Yu and Baker, Bowen and Bao, Haiming and others},
  journal={arXiv preprint arXiv:2508.10925},
  year={2025}
}

@article{2025Kimi-K2,
  title={Kimi k2: Open agentic intelligence},
  author={Team, Kimi and Bai, Yifan and Bao, Yiping and Chen, Guanduo and Chen, Jiahao and Chen, Ningxin and Chen, Ruijue and Chen, Yanru and Chen, Yuankun and Chen, Yutian and others},
  journal={arXiv preprint arXiv:2507.20534},
  year={2025}
}

@article{2024DeepSeek-V3,
  title={Deepseek-v3 technical report},
  author={Liu, Aixin and Feng, Bei and Xue, Bing and Wang, Bingxuan and Wu, Bochao and Lu, Chengda and Zhao, Chenggang and Deng, Chengqi and Zhang, Chenyu and Ruan, Chong and others},
  journal={arXiv preprint arXiv:2412.19437},
  year={2024}
}

@article{zuo2025ttrl,
  title={Ttrl: Test-time reinforcement learning},
  author={Zuo, Yuxin and Zhang, Kaiyan and Sheng, Li and Qu, Shang and Cui, Ganqu and Zhu, Xuekai and Li, Haozhan and Zhang, Yuchen and Long, Xinwei and Hua, Ermo and others},
  journal={Conference on Neural Information Processing Systems},
  year={2025}
}

@article{zhao2025absolute,
  title={Absolute zero: Reinforced self-play reasoning with zero data},
  author={Zhao, Andrew and Wu, Yiran and Yue, Yang and Wu, Tong and Xu, Quentin and Lin, Matthieu and Wang, Shenzhi and Wu, Qingyun and Zheng, Zilong and Huang, Gao},
  journal={arXiv preprint arXiv:2505.03335},
  year={2025}
}

@article{zhao2025learning,
  title={Learning to reason without external rewards},
  author={Zhao, Xuandong and Kang, Zhewei and Feng, Aosong and Levine, Sergey and Song, Dawn},
  journal={arXiv preprint arXiv:2505.19590},
  year={2025}
}

@article{wang2020tent,
  title={Tent: Fully test-time adaptation by entropy minimization},
  author={Wang, Dequan and Shelhamer, Evan and Liu, Shaoteng and Olshausen, Bruno and Darrell, Trevor},
  journal={International Conference on Representation Learning},
  year={2020}
}

@article{zhang2025right,
  title={Right Question is Already Half the Answer: Fully Unsupervised LLM Reasoning Incentivization},
  author={Zhang, Qingyang and Wu, Haitao and Zhang, Changqing and Zhao, Peilin and Bian, Yatao},
  journal={Advances in neural information processing systems},
  year={2025}
}

@inproceedings{sun2020test,
  title={Test-time training with self-supervision for generalization under distribution shifts},
  author={Sun, Yu and Wang, Xiaolong and Liu, Zhuang and Miller, John and Efros, Alexei and Hardt, Moritz},
  booktitle={International conference on machine learning},
  pages={9229--9248},
  year={2020},
  organization={PMLR}
}
\bibliographystyle{plainnat}

\newpage
\appendix
\section{Appendix}

\subsection{Test-time Reinforcement Learning}

Test-time training (TTT)~\citep{sun2020test,wang2020tent} has received increasing research interest. These approaches adapt model parameters at test time by exploiting the structure and distributional properties of incoming test data. Recent works explore self-supervised RL with various internal reward signals including majority voting~\citep{zuo2025ttrl}, semantic coherence~\citep{zhang2025right}, token-level confidence~\citep{zhao2025learning} and self-play~\citep{zhao2025absolute}. By updating models at test time using RL, test-time reinforcement learning (TTRL)~\citep{zuo2025ttrl} emerges as a promising way to enables the model to explore and improve its performance on unlabeled test set without explicit supervision. In this work, we adopt TTRL with P1-30B-A3B for further improvement due to its representativeness and effectiveness. 
Specifically, we merge all unlabeled test data of HiPhO, sampling 32 responses per sample. The majority voting consensus (cons@32) is then used as pseudo-labels to fine-tune the P1-30B-A3B model checkpoint for 64 steps with GSPO. We employ a batch size of 256, a mini-batch size of 32, and a maximum response length of 48K. Before training, we filter out multiple-choice questions via regular expression matching. In our preliminary experiments, we observed that multiple-choice questions tend to inflate the majority consensus ratio, which often leads to reward hacking. The validation results (mean@16) are shown in Table~\ref{tab:hipho_ttrl}.

\begin{table}[H]
\centering
\caption{Performance of our P1-30B-A3B model combined with TTRL on the HiPhO benchmark (note that results presented here are evaluated only in answer level, leveraging Qwen3-30B-A3B-Instruct as judge).}
\label{tab:hipho_ttrl}
\small
\resizebox{\textwidth}{!}{%
\setlength{\tabcolsep}{2.3pt}
\begin{tabular}{lccccccccccccc|c}
\toprule
\textbf{Physics Olympiad} & \multicolumn{2}{c}{\textbf{IPhO}} & \textbf{APhO} & \multicolumn{2}{c}{\textbf{EuPhO}} & \multicolumn{2}{c}{\textbf{NBPhO}} & \multicolumn{6}{c|}{\textbf{PanPhO}~~
\textbf{\footnotesize PanMechanics} ~~ \textbf{F=MA}} & \textbf{Avg.} \\
\textbf{Year} &2025&2024&2025&2025&2024&2025&2024&2025&2024&2025&2024&2025&2024& \\ \midrule
% Full Mark (Human) & 30.0 & 30.0 & 30.0 & 30.0 & 30.0 & 72.0 & 72.0 & 100.0 & 100.0 & 100.0 & 100.0 & 25.0 & 25.0 & 57.2 \\
% Full Mark (Model) & 29.4 & 29.3 & 30.0 & 29.0 & 28.0 & 43.5 & 50.0 & 100.0 & 98.0 & 100.0 & 100.0 & 25.0 & 25.0 & 52.9 \\
% Top-1 Score (Human) & 29.2 & 29.4 & 30.0 & 27.0 & 30.0 & 53.2 & 40.8 & 81.0 & 66.5 & 62.0 & 51.0 & 25.0 & 24.0 & 42.2 \\
P1-30B-A3B & 12.0 & 14.4 & 23.1 & 0.5 & 12.1 & 24.5 & 13.8 & 41.7 & 49.5 & 58.1 & 73.1 & 17.9 & 17.8 & 27.6 \\
P1-30B-A3B + TTRL & 13.8 & 15.4 & 23.6 & 0.3 & 14.1 & 24.4 & 12.8 & 42.9 & 50.4 & 61.1 & 75.2 & 18.0 & 17.8 & 28.4 \\
\bottomrule
\end{tabular}%
}
\end{table}

% \subsection{More Evaluation Results}

\subsection{Case Study}

This case study examines a complex problem from the 2025 International Physics Olympiad (IPhO), which investigates the physical principles of Cox's 18th-century timepiece—an ingenious device that harnesses atmospheric pressure fluctuations to generate energy. The problem requires determining optimal parameters to maximize energy dissipation through friction, involving multi-step reasoning that combines mechanical analysis, constraint formulation, and calculus-based optimization.

The P1-235B-A22B model achieved a perfect score (1.0/1.0) on this problem, demonstrating strong capabilities in:
\begin{itemize}[noitemsep]
    \item \textbf{Physical intuition:} Correctly identifying the critical force balance constraint at the stop position
    \item \textbf{Mathematical modeling:} Establishing proper relationships between system parameters
    \item \textbf{Optimization techniques:} Successfully applying constrained optimization to find extrema
\end{itemize}

This performance showcases P1's proficiency in handling competition-level physics problems that demand deep physical understanding and rigorous analytical reasoning.

\begin{tcolorbox}[
    enhanced,
    colback=blue!5!white,
    colframe=blue!75!black,
    title=IPhO 2025 Question 3-Part 3-C4,
    fonttitle=\bfseries\large,
    breakable
]

\textbf{\large Background:}

In 1765, British clockmaker James Cox invented a clock whose only source of energy is the fluctuations in atmospheric pressure. Cox's clock used two vessels containing mercury. Changes in atmospheric pressure caused mercury to move between the vessels, and the two vessels to move relative to each other. This movement acted as an energy source for the actual clock.

\vspace{0.4cm}

\textbf{\large Simplified Model:}

The simplified model consists of:
\begin{itemize}[noitemsep]
    \item A cylindrical bottom cistern containing a mercury bath
    \item A two-part barometric tube (completely emptied of air) dipped into the bath
    \item The cistern and tube are each suspended by cables through ideal pulleys, attached to a mass $M$ that can slide on a horizontal surface
    \item Total mercury volume: $V_{\ell} = 5$ L
\end{itemize}

\vspace{0.2cm}

The system experiences solid friction between mass $M$ and the horizontal support with magnitude $F_{\mathrm{s}}$. Two stops limit the displacement: $-X \leq x \leq X$ (where $X > 0$).

\vspace{0.2cm}

\textbf{Key parameters are defined as:}
\begin{equation}
\xi = \frac{S_{\mathrm{b}} + S_{\mathrm{c}}}{S_{\mathrm{b}} S_{\mathrm{c}}} \frac{F_{\mathrm{s}}}{A} \quad \text{and} \quad \lambda = \frac{2 \rho g X}{A}
\end{equation}

where $S_t \ll S_b, S_c$ is assumed.

\tcblower

\textbf{\large Question:}

Considering $S_{\mathrm{b}} \simeq S_{\mathrm{c}}$ and $S_{\mathrm{t}} \ll S_{\mathrm{b}}$, determine the expressions for:
\begin{enumerate}
    \item The optimal friction force $F_{\mathrm{s}}^{\star}$ as a function of $\rho, g, S_{\mathrm{c}}$ and $A$
    \item The optimal displacement $X^{\star}$ as a function of $\rho, g, S_{\mathrm{c}}$ and $A$
    \item The corresponding maximum energy $W^{\star}$
    \item Calculate the numerical value in mJ with $A = 5 \times 10^{2}$ Pa
\end{enumerate}

\vspace{0.2cm}

\textbf{Given data:} $S_c = 210$ cm$^2$, $A = 5 \times 10^2$ Pa, $g = 9.8$ m$\cdot$s$^{-2}$, $\rho_{\text{Hg}} = 13534$ kg/m$^3$
\end{tcolorbox}

\begin{tcolorbox}[
    enhanced,
    colback=violet!5!white,
    colframe=violet!75!black,
    title=P1 Solution Overview,
    fonttitle=\bfseries,
    breakable
]
To determine the optimal parameters for maximizing the energy dissipated by solid friction in Cox's timepiece, we analyze the system under the assumption that $S_b \approx S_c$ and $S_t \ll S_b, S_c$. The energy dissipated per period $\tau_1$ is given by:

\begin{equation}
W = 4 F_s X
\end{equation}

This expression arises from the fact that the mass $M$ moves between $-X$ and $X$, covering a total distance of $4X$ per cycle. The constraint on the system comes from the pressure fluctuation $P_1(t)$ with amplitude $A$, which must be sufficient to overcome both the restoring force due to mercury displacement and the friction force.
\end{tcolorbox}

\vspace{0.3cm}

\begin{tcolorbox}[
    enhanced,
    colback=violet!5!white,
    colframe=violet!75!black,
    title={(1) Optimal Friction Force $F_s^*$},
    fonttitle=\bfseries,
    breakable
]
The maximum force provided by atmospheric pressure fluctuations is $S_c A$, where $S_c$ is the cross-sectional area of the cistern. This force must balance the sum of the restoring force and the friction force at the stop position $x = X$:

\begin{equation}
S_c A = 2 S_c \rho g X + F_s
\end{equation}

Solving for $F_s$, we get:

\begin{equation}
F_s = S_c (A - 2 \rho g X)
\end{equation}

Substituting into the energy expression $W = 4 F_s X$, we obtain:

\begin{equation}
W = 4 S_c (A - 2 \rho g X) X
\end{equation}

To maximize $W$, take the derivative with respect to $X$ and set it to zero:

\begin{equation}
\frac{dW}{dX} = 4 S_c (A - 4 \rho g X) = 0 \quad \Rightarrow \quad X^* = \frac{A}{4 \rho g}
\end{equation}

Substitute back to find $F_s^*$:

\begin{equation}
F_s^* = S_c \left( A - 2 \rho g \cdot \frac{A}{4 \rho g} \right) = S_c \left( A - \frac{A}{2} \right) = \frac{S_c A}{2}
\end{equation}

\tcblower
\textbf{Answer:} \quad $\boxed{F_s^* = \dfrac{S_c A}{2}}$
\end{tcolorbox}

\vspace{0.3cm}

\begin{tcolorbox}[
    enhanced,
    colback=violet!5!white,
    colframe=violet!75!black,
    title={(2) Optimal Displacement $X^*$},
    fonttitle=\bfseries,
    breakable
]
As derived in the previous step, when we maximize the work function $W = 4 S_c (A - 2 \rho g X) X$, we obtain:

\begin{equation}
X^* = \frac{A}{4 \rho g}
\end{equation}

\tcblower
\textbf{Answer:} \quad $\boxed{X^* = \dfrac{A}{4 \rho g}}$
\end{tcolorbox}

\vspace{0.3cm}

\begin{tcolorbox}[
    enhanced,
    colback=violet!5!white,
    colframe=violet!75!black,
    title={(3) Maximum Energy $W^*$},
    fonttitle=\bfseries,
    breakable
]
Using the expressions for $F_s^*$ and $X^*$:

\begin{equation}
W^* = 4 F_s^* X^* = 4 \cdot \frac{S_c A}{2} \cdot \frac{A}{4 \rho g} = \frac{S_c A^2}{2 \rho g}
\end{equation}

\tcblower
\textbf{Answer:} \quad $\boxed{W^* = \dfrac{S_c A^2}{2 \rho g}}$
\end{tcolorbox}

\vspace{0.3cm}

\begin{tcolorbox}[
    enhanced,
    colback=violet!5!white,
    colframe=violet!75!black,
    title={(4) Numerical Value of $W^*$},
    fonttitle=\bfseries,
    breakable
]
\textbf{Given:}
\begin{itemize}[noitemsep]
    \item $S_c = 210$ cm$^2 = 0.021$ m$^2$
    \item $A = 500$ Pa
    \item $\rho = 13534$ kg/m$^3$ (density of mercury)
    \item $g = 9.8$ m/s$^2$
\end{itemize}

\vspace{0.2cm}

\textbf{Compute:}

\begin{equation}
W^* = \frac{0.021 \times (500)^2}{2 \times 13534 \times 9.8} = \frac{5250}{265266.4} \approx 0.01979\text{ J} = 19.79\text{ mJ}
\end{equation}

Rounded to two significant figures: $W^* \approx 20$ mJ

\tcblower
\textbf{Answer:} \quad $\boxed{W^* \approx 19.8\text{ mJ}}$
\end{tcolorbox}

\vspace{0.3cm}

% \begin{tcolorbox}[
%     enhanced,
%     colback=purple!10!white,
%     colframe=purple!75!black,
%     title=Summary: Final Answers,
%     fonttitle=\bfseries\Large,
%     breakable
% ]
% \begin{align}
% \text{(1)} \quad & \boxed{F_{\mathrm{s}}^{\star} = \dfrac{S_{\mathrm{c}} A}{2}} \\[0.3cm]
% \text{(2)} \quad & \boxed{X^{\star} = \dfrac{A}{4 \rho g}} \\[0.3cm]
% \text{(3)} \quad & \boxed{W^{\star} = \dfrac{S_{\mathrm{c}} A^{2}}{2 \rho g}} \\[0.3cm]
% \text{(4)} \quad & \boxed{19.8\text{ mJ}}
% \end{align}
% \end{tcolorbox}

\end{document}